\documentclass[lettersize,journal]{IEEEtran}
\usepackage{amsmath,amsfonts}
\usepackage{array}
\usepackage[caption=false,font=normalsize,labelfont=sf,textfont=sf]{subfig}
\usepackage{textcomp}
\usepackage{stfloats}
\usepackage{url}
\usepackage{mathtools}
\usepackage{verbatim}
\usepackage{multirow}
\usepackage{graphicx}
\usepackage{booktabs}
\usepackage{colortbl}
\usepackage{xcolor}
\usepackage{hyperref}
\definecolor{V}{RGB}{21,137,139}
\definecolor{X}{RGB}{234,120,60}
\usepackage{cite}
\usepackage{algorithm}
\usepackage{algpseudocode}
\usepackage[algo2e]{algorithm2e}
\usepackage[utf8]{inputenc}
\usepackage{braket}
\usepackage{amsmath,amssymb,amsfonts}%
\usepackage{xcolor}
\usepackage{hyperref}
\definecolor{V}{RGB}{21,137,139}
\definecolor{X}{RGB}{234,120,60}
\begin{document}

\title{MQAdapter: Multi-Modal Quantum Adapter for Coarse-to-Fine VLM Fine-tuning}

\author{Yumiao Zhao, Bo Jiang, Min Lu, Xiao Wang and Jin Tang 
        % <-this % stops a space
\thanks{
Yumiao Zhao, Bo Jiang, Xiao Wang, and Jin Tang are with the Anhui Provincial Key Laboratory of Multimodal Cognitive Computation, School of Computer Science and Technology, Anhui University, Hefei, China. Min Lu is with Origin Quantum Computing Company Limited, Hefei, Anhui 230026, China.

Corresponding author: Bo Jiang and Xiao Wang

%This paper was produced by the IEEE Publication Technology Group. They are in Piscataway, NJ.
}% <-this % stops a space
\thanks{Manuscript received April 19, 2021; revised August 16, 2021.}}

\markboth{Journal of \LaTeX\ Class Files,~Vol.~14, No.~8, August~2021}%
{Shell \MakeLowercase{\textit{et al.}}: A Sample Article Using IEEEtran.cls for IEEE Journals}

\IEEEpubid{0000--0000/00\$00.00~\copyright~2021 IEEE}
\maketitle
\begin{abstract}
Large-scale Vision–Language Models (VLMs) have demonstrated impressive transfer learning capabilities across a wide range of tasks. 
For few-shot classification tasks, we observe that VLMs exhibit a notable capability in candidate category filtering, and thus obtain a high Top‑$K$ accuracy.
However, they often struggle with fine-grained discrimination among visual similar categories, resulting in unsatisfactory Top‑1 performance, as shown in Figure 1. 
%observe that frozen CLIP often retrieves the correct class within the Top-k predictions, but fails to rank it first among highly similar category candidates. 
Existing studies on VLM adapters generally focus on global alignment between visual and textual representations in feature learning space, but fail to exploit similar semantic categories for fine-grained visual representation refinement. 
Based on these observations, 
we propose a novel coarse-to-fine VLM fine-tuning approach for few-shot learning by leveraging quantum computation, termed Multi-Modal Quantum Adapter (MQAdapter). 
Specifically, 
MQAdapter begins by retrieving the  Top-$K$ category candidates most similar to the input image as semantic anchors. 
Then, it develops a cross-modal quantum learning mechanism to refine visual features under the guidance of these anchors. 
The core of our cross-modal quantum learning is to encode visual and textual features into quantum states. By leveraging quantum entanglement and superposition in a high-dimensional Hilbert space, it effectively models higher-order cross-modal interactions, resulting in more discriminative representations than those from traditional Euclidean adapters.  
Note that MQAdapter is parameter-efficient and can be combined with various existing fine-tuning algorithms to achieve further performance gains. Evaluations on 15 datasets verify the effectiveness of MQAdapter while requiring fewer trainable parameters. 

\end{abstract}


\section{Introduction}
\IEEEPARstart {L}{arge-scale} Vision-Language Models (VLMs)~\cite{radford2021learning,yao2021filip,tschannen2025siglip} have become widely used foundation models for effective transfer learning in various downstream tasks. A representative VLM is CLIP~\cite{radford2021learning}, which is trained on 400 million image–text pairs and aligns the two modalities via contrastive pre-training. As a result, CLIP serves as an effective transferable backbone for various downstream tasks, including few-shot classification~\cite{li2024graphadapter,zhu2024awt}, semantic segmentation~\cite{fahes2024simple,zhao2025med}, and visual question answering~\cite{ozdemir2024enhancing,ye2023video}. However, adapting VLMs by updating all model parameters requires considerable computational resources and increases the risk of overfitting, especially in few-shot scenarios. Therefore, researchers explore parameter-efficient fine-tuning (PEFT) methods for adapting large VLMs to downstream tasks with a small number of trainable parameters.

\begin{figure}[t]
  \centering
  \includegraphics[width=1\linewidth]{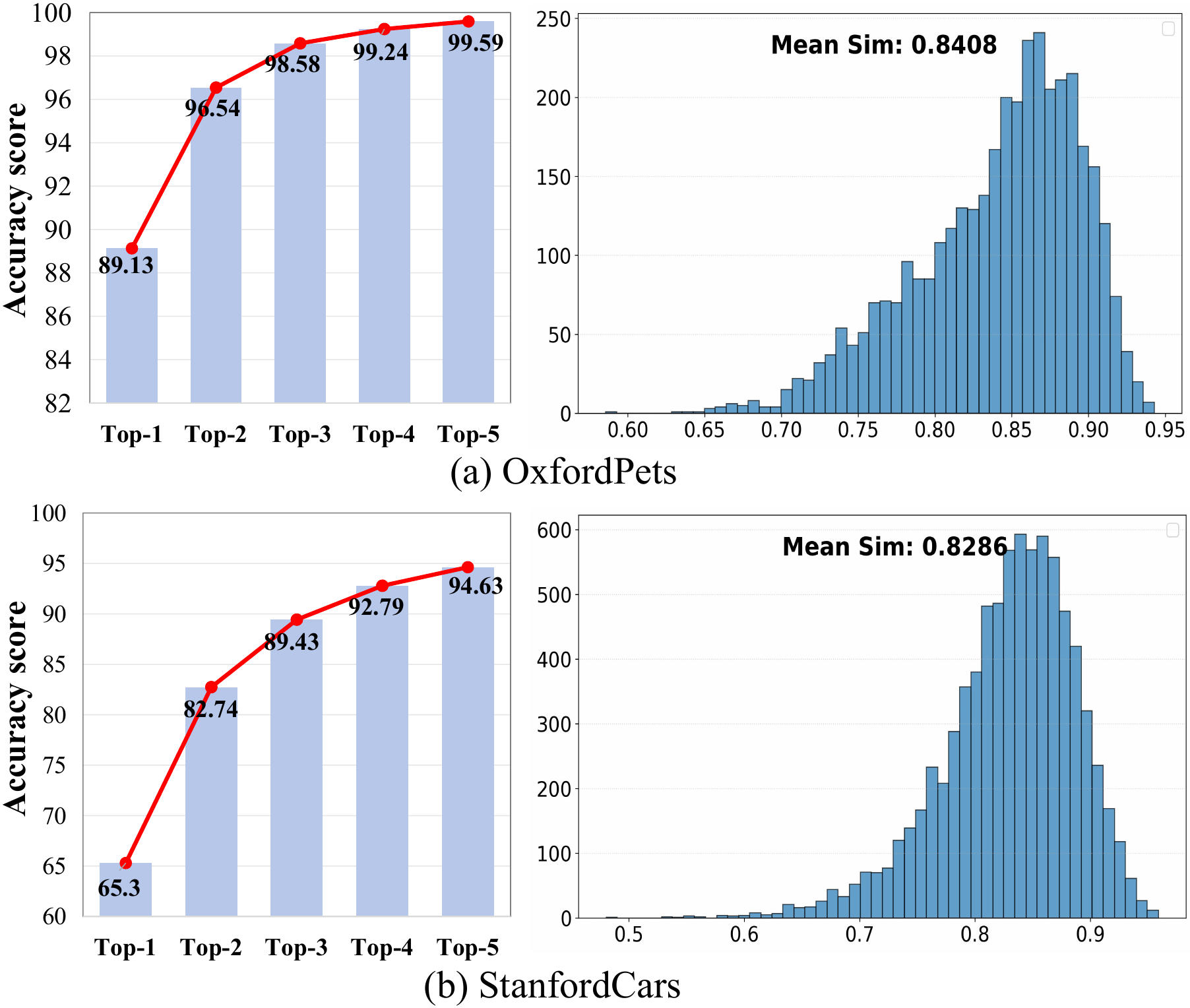}
  \caption{Performance and feature-similarity analysis of frozen CLIP on OxfordPets and StanfordCars datasets. Left: Top-1 to Top-5 accuracy on the test dataset. Right: 
  Distribution of similarities between each input image feature and its Top-$5$ predicted categories.}
\label{v_motivation}
\end{figure}

\IEEEpubidadjcol 

Existing few-shot VLM adaptation approaches include prompt learning and adapter-based methods. Prompt learning methods typically introduce learnable prompts to adapt VLMs to target tasks. For instance, CoOp~\cite{zhou2022learning}, CoCoOp~\cite{zhou2022conditional}, and ATPrompt~\cite{li2025atprompt} replace hand-crafted textual templates with learnable continuous context tokens or incorporate attribute-aware words into the text prompts to facilitate task adaptation. Beyond input-level prompt tuning, some methods such as MaPLe~\cite{khattak2023maple}, PromptSRC~\cite{khattak2023self}, and HiCroPL~\cite{zheng2025hierarchical} integrate learnable tokens into intermediate layers of pre-trained VLMs, enabling the models to capture task-specific knowledge while maintaining cross-modal alignment. Adapter-based methods aim to design a lightweight module that are inserted into pre-trained VLMs to enable parameter-efficient adaptation. For example, GraphAdapter~\cite{li2023graphadapter} and Hegraphadapter~\cite{zhao2024hegraphadapter} leverage graph convolutional networks to capture structured cross-modal relationships, while CLIP-Adapter~\cite{gao2024clip} and MMA~\cite{zhang2024multi} introduce feature transformation modules that operate on visual and/or textual representations, enhancing visual-text alignment. 

In this paper, we first observe that, for few-shot learning tasks, pre-trained VLMs can exhibit a notable capability in filtering candidate categories and thus obtain a high Top‑$K$ accuracy.
Nevertheless, they often struggle with fine-grained discrimination among visual similar categories, resulting in unsatisfactory Top‑1 performance.
Figure~\ref{v_motivation} shows an example: while Top-$1$ accuracy is  low, Top-$5$ accuracy exceeds $90\%$. 
A similar observation has also been obtained in a recent study~\cite{yang2026endowing}. 
 \emph{This suggests that pre-trained VLMs tend to encode only \textbf{coarse} representations for visual data. Therefore, how to enhance these representations with \textbf{fine-grained} discrimination becomes crucial when adapting VLMs to downstream tasks.}   
However, after reviewing the above studies on VLM fine-tuning adapters, we observe that existing adapter approaches primarily focus on global alignment between visual and textual representations in traditional feature spaces. They generally overlook fine-grained cues among similar categories, thereby limiting their effectiveness in few-shot scenarios.

Based on the above observations, 
this paper proposes a novel coarse-to-fine VLM fine-tuning approach for few-shot learning by 
leveraging quantum computation theory~\cite{biamonte2017quantum,huang2021power,hur2022quantum, benedetti2019parameterized}, termed Multi-Modal Quantum Adapter (MQAdapter). 
%is known to enhance representational capacity in complex learning systems by exploiting superposition and entanglement~\cite{pokharel2025quantum}. 
 MQAdapter begins by retrieving the Top-$K$ textual categories most similar to the input visual image ${I}$ via pre-trained VLM model, which serve as \emph{semantic anchors} $\mathcal{P}$ to guide the adaptation process.
Then, the subsequent goal of our MQAdapter is to learn a highly discriminative visual representation for image ${I}$ by fully exploiting the modality interactions between ${I}$ and its semantic anchors $\mathcal{P}$.
%It is known that quantum computation can enhance representational capacity in complex learning systems by exploiting superposition and entanglement~\cite{pokharel2025quantum}.
To achieve this, inspired by the superposition and entanglement mechanism~\cite{pokharel2025quantum} in quantum computation field, we equip MQAdapter with a novel \textbf{cross-modal quantum learning} mechanism to enhance the representation of visual features under the guidance of semantic anchors. Specifically, it first encodes visual and textual features into a unified quantum representation space using a shared variational quantum circuit (VQC). 
Then, it models higher-order cross-modal interaction effectively by leveraging quantum entanglement and superposition within a high-dimensional Hilbert space~\cite{2002Quantum}, resulting in more discriminative representations than those from traditional Euclidean adapters. 
Finally, the quantum-enhanced visual features are projected back to the original feature dimension and injected into the frozen visual representation to achieve adaptation. 
%We insert CQA into multiple CLIP layers, enabling MQAdapter toprogressively refine visual features for effective downstream adaptation.
MQAdapter is parameter-efficient and can be combined with various existing fine-tuning approaches to achieve fine-grained VLM adaptations.  

We note that the quantum learning models have been employed some large-scale foundation model fine-tuning tasks in recent years~\cite{koike2025quantum,kong2025quantum,liu2024quantum}. 
These works generally utilize quantum or quantum-inspired parameterizations to enable compact full-rank or high-rank adaptation for LLMs.
Differently, we focus on multi-modal VLM fine-tuning. 
The key difference of our MQAdapter is to leverage  parameterized quantum circuits 
to achieve \textbf{fine-grained  
modality-interaction} between visual and textual semantic anchors, which ultimately enhances visual-text fine-grained alignment for VLM adaptation on downstream tasks. 
To our best knowledge, this is the first work to explore quantum computation for multi-modality adapter.

Overall, the main contributions of this paper are summarized as follows:
\begin{itemize}
    \item We propose a novel coarse-to-fine VLM fine-tuning approach, termed MQAdapter by leveraging 
quantum computing model. 
 %   Multi-modal Quantum Adaptation (MQAdapter) framework. 
    To the best of our knowledge, this is the first work to explore multi-modal quantum computation for parameter-efficient few-shot VLM adaptation. 
    
    \item We design a cross-modal quantum learning model to capture high-order interactions between visual and textual modalities in a shared quantum Hilbert space, resulting in fine-grained image-text representation learning.
    \item MQAdapter follows a plug-and-play design for VLM adaptation. It can be integrated into different VLM fine-tuning frameworks and consistently improves performance while requiring only 0.078M additional trainable parameters. 
\end{itemize}

The remainder of this paper is organized as follows. In Section II, we review the related work. Section III introduces the preliminaries of quantum computing. Section IV presents the proposed MQAdapter. Section V reports experiments and ablation studies. Section VI concludes the paper.

\section{Related Work}
\label{relatedwork}

\subsection{Efficient Transfer Learning for VLMs}
Large-scale Vision-Language Models (VLMs) are pre-trained on large-scale image–text pairs and demonstrate strong generalization abilities across various downstream applications~\cite{zhong2022regionclip,basak2025semidavil,fahes2024simple,zhao2025med,kim2024vlm}. VLMs such as CLIP~\cite{radford2021learning}, FILIP~\cite{yao2021filip}, and SigLIP2~\cite{tschannen2025siglip} learn a shared embedding space through contrastive learning, enabling effective zero-shot and few-shot transfer. Despite their success, directly adapting these models to downstream tasks remains challenging. To address these challenges, recent works are focused on Parameter-Efficient Transfer Learning (PETL) methods that adapt pretrained VLMs to downstream tasks while maintaining their strong generalization ability. PETL methods can be broadly categorized into prompt learning and adapter-based approaches. 

Prompt learning~\cite{khattak2023maple,khattak2023self,Conceptual2024,hu2024comma} optimizes a limited set of tokens for adapting pretrained VLMs. For example, CoOp~\cite{zhou2022learning} introduces learnable context tokens to construct task-specific text prompts, replacing hand-crafted templates. 
CoCoOp~\cite{zhou2022conditional} further refines the textual representation by conditioning the prompts on visual features from each input sample. TextRefiner~\cite{xie2024textrefiner} enriches text prompts with local visual concepts extracted from the image branch of VLMs. 
% DPC~\cite{li2025dpc} adopts a weighting-decoupling strategy to balance task-specific adaptation and generalization to unseen classes.
ATPrompt~\cite{li2025atprompt} enriches textual templates with class-specific attribute descriptions, thereby enhancing the transferability of prompt learning.
TCP~\cite{yao2024tcp} enhances generalization by transforming textual priors into class-aware prompt tokens.
% MMRL~\cite{guo2025mmrl} introduces a shared latent representation that is projected into both image and text tokens, enabling richer multi-modal interactions during adaptation.

Adapter-based methods~\cite{zhao2025finegrained,zhang2024multi,li2023graphadapter} introduce lightweight modules into pretrained VLMs to facilitate task-specific adaptation. For example, CLIP-Adapter~\cite{gao2024clip} introduces two-layer MLP adapters in the visual branch to learn task-specific feature representations. Tip-Adapter~\cite{zhang2022tip} adopts a training-free strategy by caching few-shot visual embeddings and performing similarity-based inference at test time. However, these methods typically adapt visual and textual modalities independently. To enable cross-modal interaction, MMA~\cite{zhang2024multi} uses a shared fully connected layer to establish connection between encoders, allowing gradients to propagate across branches and improving alignment. PVA~\cite{lu2025variational} introduces modality-specific variational adapters attached to align visual and textual representations at the distribution level, enabling robust adaptation under data imbalance.

\subsection{Quantum Machine Learning}
Quantum Machine Learning (QML)~\cite{biamonte2017quantum,huang2021power,hur2022quantum} introduces a novel learning paradigm that integrates quantum computing with classical machine learning methods. QML approaches employ variational quantum circuits, leveraging quantum entanglement and superposition to model complex dependencies within data. 
Recently, QML has also been explored for efficient adaptation of Large Language Models (LLMs). For example, Quantum-PEFT~\cite{koike2025quantum} leverages parameterized quantum circuits to realize full-rank and parameter-efficient LLM adaptation. QTHA~\cite{kong2025quantum} integrates quantum neural networks as an adaptation module to alleviate the limitations of classical low-rank fine-tuning. Similarly, Quantum Parameter Adaptation (QPA)~\cite{liu2024quantum} adopts hybrid quantum-classical mappings to generate compact fine-tuning parameters. QuanTA~\cite{chen2024quanta} is inspired by quantum circuits and uses a unitary matrix and tensor parameterizations to enable high-rank LLM fine-tuning. These methods aim to overcome the expressiveness limitations of conventional low-rank fine-tuning, enabling compact full-rank or high-rank adaptation for LLMs. Beyond language models, QML has also been investigated in vision and multi-modal learning. 
For example, Zhang et al.~\cite{zhang2023quantum} simulate quantum computation by representing hyperspectral cuboids in a quantum-like form with amplitude and phase, thereby modeling spectral–spatial interactions for hyperspectral image classification. Pokharel et al.~\cite{pokharel2025quantum} leverage quantum entanglement for multi-modal fusion and introduce a modality-aware circuit isolation mechanism to ensure stable quantum training under missing modalities. QViLa~\cite{mukesh2024qvila} first combines extracted visual and textual embeddings and then employs a quantum-augmented layer to enhance global multi-modal feature fusion. MEDQUA~\cite{li2026medqua} inserts a quantum adapter into pretrained medical VLM decoders, where selected decoder tokens are routed to a shallow quantum module and all tokens are processed by a LoRA-like classical path.
 Unlike previous works, our MQAdapter fully leverages multi-modal cues to guide visual feature adaptation for VLMs, where different modalities are mapped into a unified quantum-enhanced representation space to model fine-grained cross-modal interactions and generate adaptation biases for frozen VLM layers.

\begin{figure*}[htbp]
  \centering
  \includegraphics[width=0.9\linewidth]{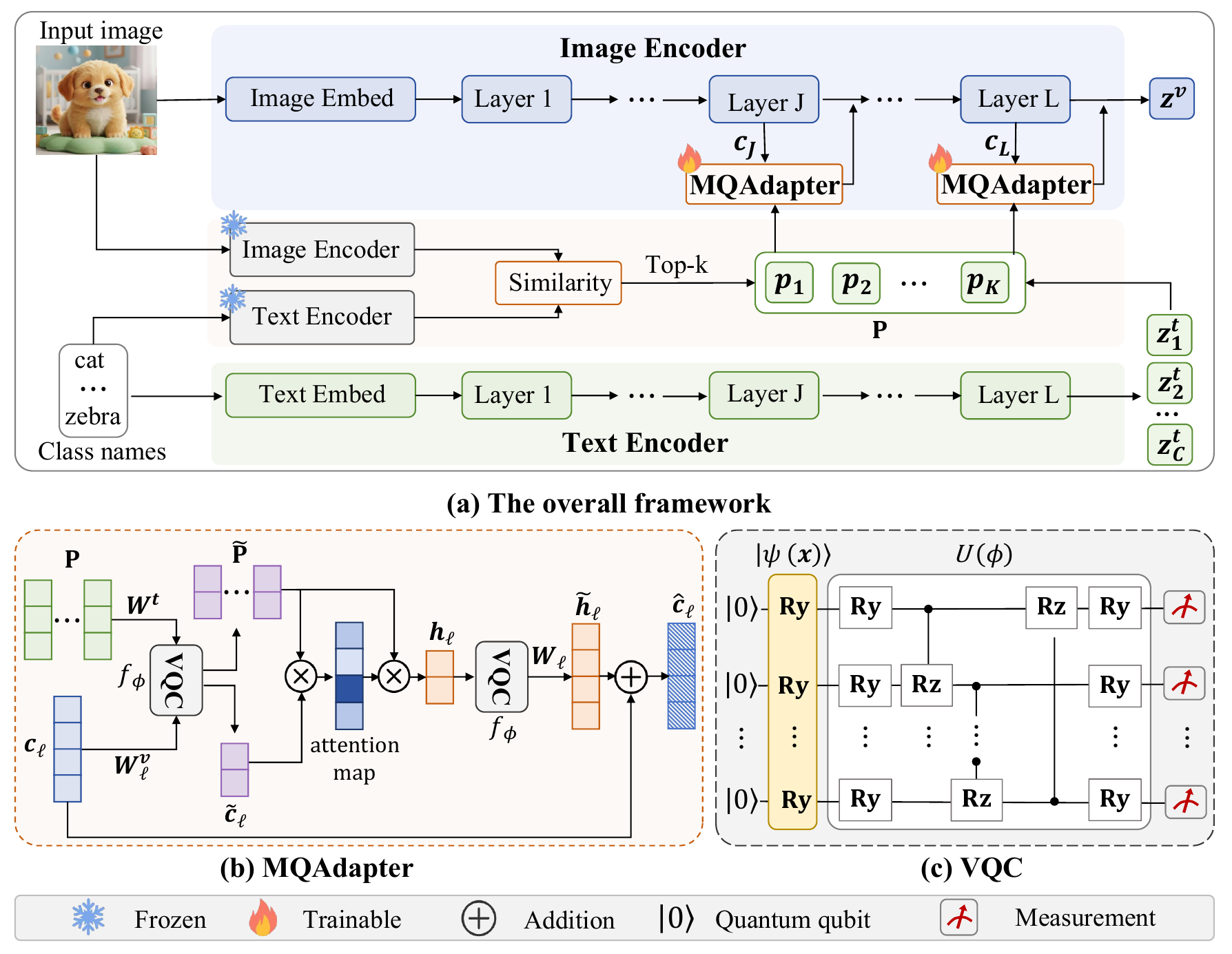}
  \caption{ The overall framework of the proposed coarse-to-fine Multi-modal Quantum Adapter (MQAdapter). In the coarse stage, we leverage a frozen CLIP model to retrieve the Top-$K$ most relevant text categories as semantic anchors. In the fine stage, these anchors guide the refinement of the image encoder through the cross-modal quantum learning  mechanism, which captures high-order interactions between modalities in quantum representation space, thereby producing more discriminative visual representations and improving few-shot adaptation performance on downstream tasks.
  } 
  
\label{network}
\end{figure*}

\section{Preliminaries of Quantum Computing}
\label{Preliminaries}
\subsection{Quantum States and Superposition}
The state of a quantum system is represented in a complex Hilbert space, which encodes both the amplitude and phase.
Phase information enables quantum interference, allowing different components of a quantum state to combine constructively or destructively~\cite{2002Quantum}. An $N$-qubit system is represented in a $2^N$-dimensional complex Hilbert space $\mathbb{C}^{2^N}$. Quantum states are commonly represented using Dirac notation~\cite{Dirac_2008rpq}. The state of a single qubit can be written as:
\begin{equation}
    |\psi\rangle = \alpha |0\rangle + \beta |1\rangle,
\end{equation}
where $\alpha, \beta$ are the amplitudes and satisfy $|\alpha|^2 + |\beta|^2 = 1$. The $|0\rangle$ and $|1\rangle$ denote the computational basis of Hilbert space. The probabilities of observing the states $|0\rangle$ or $|1\rangle$ are given by $|\alpha|^2$ and $|\beta|^2$, respectively. 
In addition, the qubit state can be visualized on the Bloch sphere which is rewritten as:
\begin{equation}
    |\psi\rangle = \cos(\frac{\theta}{2})|0\rangle + e^{i\varphi}\sin(\frac{\theta}{2})|1\rangle,
\end{equation}
where $\theta \in [0,\pi]$ and $\varphi \in [0,2\pi)$. The $i$ represents the imaginary number and satisfies $i^2=-1$. Unlike classical bits, which are restricted to binary states, qubits can exist in superpositions of $|0\rangle$ and $|1\rangle$, thereby 
allowing more expressive representations with compact parameterization.
\subsection{Parameterized Quantum Circuit}
Quantum gates serve as the fundamental building blocks of quantum computation. Parameterized rotation gates adjust the orientation of individual qubits, and controlled gates establish correlations and entanglement between qubits. 
A sequence of parameterized quantum gates forms a quantum circuit. Using a classical optimizer~\cite{wen2024evolution}, the parameters of the quantum circuit are updated to approximate a desired quantum state. Formally, it can be represented as a parameterized unitary transformation 
$U(\boldsymbol{\phi})$ acting on an $n$-qubit system as:
\begin{equation}
|\psi(\boldsymbol{\phi})\rangle = U(\boldsymbol{\phi}) |0\rangle^{\otimes N},
\end{equation}
where $\boldsymbol{\phi}$ denotes the set of trainable parameters associated with the quantum gates.
\subsection{Quantum Measurement}
Quantum measurement aims to extract classical representations from a quantum state~\cite{tilly2022variational}. Given an $n$-qubit state $|\psi(\boldsymbol{\phi})\rangle$ output from the quantum circuit, the expectation value associated with an observable $O_m$ is given by:
\begin{equation}
\boldsymbol{q}_m(\boldsymbol{\phi}) = \langle \psi(\boldsymbol{\phi}) | O_m | \psi(\boldsymbol{\phi})\rangle,
\end{equation}
In quantum computation, classical data are first encoded into quantum states. A quantum circuit is then applied to transform these states. Finally, the quantum outputs are decoded into classical representations through measurement.

%\section{Approach}
\section{The Proposed MQAdapter}
Following previous studies~\cite{gao2024clip,guo2025mmrl,li2025atprompt,li2025dpc}, we build on the pre-trained CLIP model~\cite{radford2021learning}. 
MQAdapter first retrieves Top-$K$ candidate categories as semantic anchors using a frozen pre-trained CLIP model. 
Then, it models high-order cross-modal interaction between visual features and semantic anchors via quantum learning, thereby refining visual representations and enhancing few-shot downstream adaptation. 
In this section, we introduce these processes in detail. 
%Importantly, MQAdapter is plug-and-play and can be combined with existing parameter-efficient fine-tuning frameworks.

\subsection{Semantic Anchors}
The pre-trained VLM model (e.g., CLIP) consists of a visual encoder and a text encoder. 
%We denote the \emph{frozen} pre-trained CLIP encoders as ${\mathcal{V}}^*(\cdot)$ and ${\mathcal{T}}^*(\cdot)$, respectively. 
%For the downstream task, the descriptions of the categories of text are indicated as $\{\mathbf{I}^t_1, \cdots,\mathbf{I}^t_C\}$, where $C$ is the number of classes.
For few-shot learning tasks, given an input visual image ${I}$ and all textual categories $\{{t}_1, {t}_2\cdots {t}_C\}$, where $C$ denotes the number of classes, we first obtain the feature descriptors for them via the \emph{frozen} pre-trained CLIP visual and text encoder, respectively.  
We compute the similarities between image ${I}$ and all $C$ text categories using their feature descriptors, and then select Top-$K$ candidate categories for image ${I}$  with the highest similarity scores. 
In the following, we denote 
$$
\mathbf{P}=[\boldsymbol{p}_1,\boldsymbol{p}_2\cdots \boldsymbol{p}_K]
$$ 
as the features of top $K$ candidate categories, and refer to them as semantic anchors. 

As illustrated in Figure 1, pre-trained  CLIP model can obtain a high Top-$K$ accuracy, exhibiting a notable capability in filtering candidate categories. However, it often struggles with fine-grained
discrimination among visual similar categories, resulting in
unsatisfactory Top-1 performance. 
Based on this observation, the aim of our MQAdapter is to obtain a highly discriminative visual representation by fully exploiting the relationships between visual image $I$ and its semantic anchors $\mathbf{P}$ via cross-modal quantum learning. 
In the following, we first introduce the variational quantum circuit (VQC) in our cross-modal quantum learning, and then present the adaptation process of our MQAdapter.

\subsection{Variational Quantum Circuit}
% Quantum computation leverages properties such as superposition and entanglement to represent features in a high-dimensional Hilbert space and capture non-separable correlations, enabling parameter-efficient modeling of nonlinear and higher-order interactions~\cite{kong2025quantum}. Inspired by this, 
We leverage quantum computation theory and design a Variational Quantum Circuit (VQC) function $f_{\phi}(\cdot)$ that transforms classical input features into quantum-enhanced feature representations. As shown in Fig.~\ref{network} (c), the VQC consists of three steps: data encoding, parameterized quantum circuit (PQC), and measurement. 
Given a feature vector $\boldsymbol{x}=(x_1,\dots,x_N)\in\mathbb{R}^{N}$, we first encode it into an $N$-qubit quantum state via angle encoding~\cite{weigold2020data,kim2025quantum}:
\begin{equation}
\ket{\psi(\boldsymbol{x})}=
\bigotimes_{n=1}^{N} \left(\mathrm{R_y}(x_n)\ket{0}_n\right),
\end{equation}
where $\mathrm{R_y}(\cdot)$ is a single-qubit rotation gate. 
Subsequently, we construct a parameterized quantum circuit $U(\boldsymbol{\phi})$ to transform the encoded quantum state $\ket{\psi(\boldsymbol{x})}$ as follows:
\begin{align}
U(\boldsymbol{\phi})
&=
\Big(\bigotimes_{n=1}^{N}\mathrm{R_y}(\boldsymbol{\phi}^{(2)}_{n})\Big)\cdot
\Big(\prod_{(n,m)\in\mathcal{E}}\mathrm{CRZ}(\boldsymbol{\phi}_{n,m})\Big)
\nonumber\\
&\quad\cdot
\Big(\bigotimes_{n=1}^{N}\mathrm{R_y}(\boldsymbol{\phi}^{(1)}_{n})\Big), \\
\ket{\psi'(\boldsymbol{x})}&=U(\boldsymbol{\phi})\ket{\psi(\boldsymbol{x})},
\end{align}
where $\mathcal{E}=\left\{(n,n+1)\right\}_{n=1}^{N-1}
\cup
\left\{(N,1)\right\}$ adopts a ring nearest-neighbor entanglement, and $\boldsymbol{\phi}^{(1)}$ and $\boldsymbol{\phi}^{(2)}$ denote the learnable rotation parameters. $\mathrm{CRZ}\!\left(\boldsymbol{\phi}_{n,m}\right)$ represents a two-qubit controlled-Z rotation gate~\cite{benedetti2019parameterized} that introduces entanglement between qubits $n$ and $m$. 
Finally, each qubit is measured via the Pauli-$Z$~\cite{benedetti2019parameterized} observable to produce a quantum-enhanced feature representation:
\begin{equation}
\begin{aligned}
\label{eu_vqc}
f_{\phi}(\boldsymbol{x})
&=
\big[
\langle \psi'(\boldsymbol{x})|Z_1|\psi'(\boldsymbol{x})\rangle,\dots,
\langle \psi'(\boldsymbol{x})|Z_N|\psi'(\boldsymbol{x})\rangle
\big],
\end{aligned}
\end{equation}
where $Z_N$ denotes the Pauli-$Z$ operator acting on the $N$-th qubit.

\subsection{Fine-grained Adaptation}
% In this stage, we aim to learn a highly discriminative visual representation by fully exploiting the modality interactions between the visual and its semantic anchors $\mathbf{P}^t$.
% After the coarse stage retrieves a small set of semantically similar candidates, the fine stage exploits these as semantic anchors to guide the visual encoder refinement.
% Instead of adapting the visual representation with conventional cross-attention, 
%Existing quantum-based multi-modal methods~\cite{pokharel2025quantum,wu2025multimodal,zheng2024quantum} typically encode different modalities with separate quantum registers, which may increase the quantum resources. To mitigate this issue, 

As shown in Fig.~\ref{network} (b), given 
the input image $I$ feature representation and associated semantic anchors $\mathbf{P}$, 
our MQAdapter first projects them to the same dimension and employs a shared VQC (Eq.~\ref{eu_vqc}) module to map visual features and textual semantic anchors into a unified quantum representation space. This design yields modality-consistent representations, effectively alleviating the visual-text modality gap in subsequent interactions. 
We then compute attention weights to quantify cross-modal relevance and obtain multi-modal features. These features are subsequently fed into the same shared VQC, where quantum superposition and entanglement operations enable higher-order cross-modal interactions within a high-dimensional Hilbert space.

To be specific, 
% CQL is inserted into the multiple layers of the VLM to progressively refine the visual encoder, thereby producing more discriminative representations and improving visual–text alignment for downstream adaptation.
given an input image ${I}$, we obtain the visual representations of each CLIP layer as:
\begin{align}
[\boldsymbol{c}_0,\boldsymbol{v}_0]&=\mathrm{ImageEmbed}({I}),\\
[\boldsymbol{c}_\ell,\boldsymbol{v}_\ell]&=\mathcal{V}([\boldsymbol{c}_{\ell-1},\boldsymbol{v}_{\ell-1}]) \quad \ell = 1, \ldots, L
\end{align}
where $\mathrm{ImageEmbed}(\cdot)$ splits the input image ${I}$ into fixed-size patches and projects them into feature embeddings and $\mathcal{V}(\cdot)$ denotes visual encoder. $\boldsymbol{c}_\ell\in \mathbb{R}^{d_v}$ denotes the class token at the $\ell$-th visual layer.
% This design facilitates cross-modal alignment while enabling effective high-order interactions with a compact number of qubits, thereby improving computational efficiency. 
Next, we project the visual class token $\boldsymbol{c}_\ell$ and the text semantic anchors $\mathbf{P}=[\boldsymbol{p}_1,\cdots, \boldsymbol{p}_K]$ into a $N$-dimensional space
as:
\begin{align}
\bar{\boldsymbol{c}}_\ell = \mathbf{W}^v_\ell\boldsymbol{c}_\ell ;\; \bar{\mathbf{P}} = \mathbf{W}^t\mathbf{P},
\end{align}
where $\mathbf{W}^v_\ell$ and $ \mathbf{W}^t$ denote learnable modality-specific projection matrices and $N$ denotes the number of qubits. This projection aligns the input dimensions with the number of qubits, since each input dimension is encoded as the rotation angle of one qubit.

The projected visual and textual features are encoded by using a shared VQC module $f_{\phi}(\cdot)$ (Eq.(8)):
\begin{equation}
\tilde{\boldsymbol{c}}_\ell = f_{\phi}(\bar{\boldsymbol{c}}_\ell), \quad
\tilde{\mathbf{P}}  = [f_{\phi}(\bar{\boldsymbol{p}_1}),\cdots,f_{\phi}(\bar{\boldsymbol{p}_K})].
\end{equation}
By sharing the VQC across modalities, visual and textual features are mapped into a unified quantum-enhanced representation space. Then, we compute the cross-modal attention  between the quantum-enhanced visual token and semantic anchors as:
\begin{align}
\boldsymbol{\alpha}_\ell &= \mathrm{softmax}\!\left(\frac{ \tilde{\mathbf{P}}^{\top}\tilde{\boldsymbol{c}}_\ell}{\sqrt{ N}}\right)\in \mathbb{R}^K,\\
\boldsymbol{h}_\ell& = \tilde{\mathbf{P}}\boldsymbol{\alpha}_\ell= \sum_{k=1}^{K} \alpha_{\ell,k}\,\tilde{\boldsymbol{p}}_k,
\end{align}
where $\boldsymbol{h}_\ell$ aggregates semantic priors from the selected textual anchors and obtains a multi-modal representation. 
After that, we feed $\boldsymbol{h}_\ell$ into the shared VQC to enable cross-modal higher-order interaction. Finally, the quantum-enhanced multi-modal feature is projected back to the original visual feature dimension through projection $\mathbf{W}_\ell$ 
and injected into the visual representation via a residual connection as:
\begin{align}
\hat{\boldsymbol{c}}_\ell &= \boldsymbol{c}_\ell + \gamma\,\mathbf{W}_\ell f_{\phi}(\boldsymbol{h}_\ell),
\end{align}
where $\gamma$ is a scaling hyperparameter.

\subsection{Fine-tuning for Downstream Tasks}
For the downstream task, as shown in Fig.~\ref{network} (a), we integrate MQAdapter into the last several layers of CLIP model to enable task-specific adaptation while maintaining training efficiency. Given an input image ${I}$ and a set of $C$ category-level textual descriptions $\{{t}_1, \cdots,{t}_C\}$, we obtain the refined visual representations $\boldsymbol{z}^v$ and textual representations $\{\boldsymbol{z}^t_1,\cdots,\boldsymbol{z}^t_C\}$.
The model is optimized using a temperature-scaled cross-entropy loss, which is formulated as:
\begin{equation}
\mathcal{L}_{\mathrm{cls}}
=
- \log
\frac{
\exp\!\left( \cos\!\left( \boldsymbol{z}^v, \boldsymbol{z}^t_y\right) / \tau \right)
}{
\sum_{i=1}^{C}
\exp\!\left( \cos\!\left( \boldsymbol{z}^v, \boldsymbol{z}^t_i \right) / \tau \right)
},
\end{equation}
where $y$ denotes the ground-truth label. By modeling cross-modal interactions in the quantum representation space, our framework jointly optimizes the image and text encoders, enhancing adaptation performance for downstream tasks.

\section{Experiments}
\label{experiment}
\subsection{Experimental Setup}
\subsubsection{Datasets and Tasks}
Following prior works~\cite{khattak2023maple,zhang2024multi,khattak2023self,guo2026mmrl}, we evaluate our models’ performance on four tasks, including base-to-new generalization, few-shot image classification, cross-dataset transfer, and domain generalization. All experiments are conducted under the 16-shot setting, except for the few-shot learning experiments.

% Table generated by Excel2LaTeX from sheet 'Sheet2'

\noindent \textbf{Base-to-New Generalization:} In this task, each dataset is equally split into base and new classes. The model is fine-tuned on base classes using a $16$-shot setting and is evaluated on both base and new classes. This evaluation assesses the balance between effective transfer learning on base classes and preserving the generalization when adapting to new classes. We perform these experiments on 11 image classification datasets: ImageNet~\cite{deng2009imagenet}, Caltech101~\cite{fei2004learning}, OxfordPets~\cite{parkhi2012cats}, StanfordCars~\cite{krause20133d}, Flowers102~\cite{nilsback2008automated}, Food101~\cite{bossard2014food}, FGVCAircraft~\cite{maji2013fine}, SUN397~\cite{xiao2010sun}, DTD~\cite{cimpoi2014describing}, EuroSAT~\cite{helber2019eurosat}, and UCF101~\cite{soomro2012ucf101}. 

\noindent \textbf{Few-Shot Image Classification:} This task is designed to evaluate the model’s adaptation ability under limited-data conditions. Specifically, we train the model with 1, 2, 4, 8, and 16 samples (shots) per class, and report the performance on the full test set. This evaluation is performed on the same 11 image classification datasets as the base-to-new generalization task.

\noindent \textbf{Cross-Dataset Transfer:} To assess the model’s ability to generalize to new datasets, we regard ImageNet as the source dataset and others as target datasets. After training with few-shot samples from all 1,000 ImageNet classes, the model is directly applied to the target datasets without updating any parameters.

\noindent \textbf{Domain Generalization:} To evaluate robustness to unseen domains, we train the model on ImageNet and evaluate it on four ImageNet variants with different distribution shifts, including ImageNet-V2~\cite{recht2019imagenet}, ImageNet-Sketch~\cite{wang2019learning}, ImageNet-A~\cite{hendrycks2021natural}, and ImageNet-R~\cite{hendrycks2021many}.

\subsubsection{Implementation details}
To verify the effectiveness of the proposed Multi-modal Quantum Adapter (MQAdapter), we plug MQAdapter into three representative PEFT frameworks, including MaPLe~\cite{khattak2023maple}, PromptSRC~\cite{khattak2023self}, and MMRL++~\cite{guo2026mmrl}. All experiments are conducted under a 16-shot setting, except for the few-shot learning task. We adopt the CLIP model with a ViT-B/16 backbone for all methods. For a fair comparison, we follow the training protocols of the corresponding PEFT frameworks, keeping the batch size, learning rate, and number of training epochs consistent with their original implementations. We implement the 8-qubit shared VQC using TorchQuantum. The MQAdapter is inserted into the visual encoder starting from the $6$-th transformer layer to the final layer. The adapter scaling parameter is set to $\gamma=0.001$ for the base-to-new task. For the few-shot image classification, we set  $\gamma=0.001$ for MaPLe, $\gamma=0.0001$ for PromptSRC and  $\gamma=0.002$ for MMRL++.  During the coarse stage, we select the Top-$5$ most relevant text categories as semantic anchors for guiding fine-grained visual feature representation.
 
\begin{table*}[!h]
 \caption{Base-to-new generalization performance of three representative fine-tuning methods w/ or w/o our MQAdapter on 11 datasets. MQAdapter consistently achieves the best performance on both base and new classes. Results marked with $^{*}$ are reproduced using the official implementations under the same experimental protocol, while the remaining baseline results are taken from the corresponding papers.
}\label{base_new}
    \centering
    \renewcommand{\arraystretch}{0.8}
\begin{tabular}{c|ccc|ccc|ccc|ccc} 
\toprule
\multirow{2.4}{*}{Method} & \multicolumn{3}{c|}{\textbf{Avg. over 11 datasets}} & \multicolumn{3}{c|}{ImageNet}   & \multicolumn{3}{c|}{Caltech101} & \multicolumn{3}{c}{OxfordPets}    \\ 
\cmidrule(lr){2-4}\cmidrule(lr){5-7}\cmidrule(lr){8-10}\cmidrule(lr){11-13}
                        & Base  & New   & H                            & Base  & New   & H              & Base  & New   & H              & Base  & New   & H                 \\ 
\midrule
CLIP  &69.34 &74.22 &71.70 &72.43 &68.14 &70.22 &96.84 &94.00 &95.40 &91.17 &97.26 &94.12\\
COMMA &82.42 &75.87 &79.04 &76.04 &70.89 &73.86 &97.94 &94.56 &96.50 &95.62 &97.84 &96.72\\
MMA  &83.20 &76.80 &79.87 &77.31 &71.00 &74.02 &98.40 &94.00 &96.15 &95.40 &98.07 &96.72\\
TCP &84.13 &75.36 &79.51 &77.27 &69.87 &73.38 &98.23 &94.67 &96.42 &94.67 &97.20 &95.92 \\
PromptSRC+DePT  &84.08 &75.03 &79.29 &77.91 &70.77 &74.17 &98.37 &94.14 &96.21 &94.83 &97.21 &96.00 \\
PromptSRC+DPC    &86.10 &74.78 &80.04 &78.48 &70.72 &74.40 &98.90 &94.21 &96.50 &96.13 &97.30 &96.71 \\
PromptSRC+ATP     &84.30 &76.45 &80.18 &77.69 &70.83 &74.10 &98.23 &94.91 &96.54 &95.64 &97.43 &96.53 \\
\midrule
*MaPLe                  & 81.22 & 75.08 & 78.03                        & 76.89 & 70.48 & 73.54          & 97.91 & 94.39 & 96.12          & 95.44 & 98.02 & 96.71             \\
\cellcolor{cyan!15}\textbf{+MQAdapter} & \cellcolor{cyan!15}\textbf{83.43} & \cellcolor{cyan!15}\textbf{77.40} & \cellcolor{cyan!15}\textbf{80.31}                        & \textbf{77.15} & \textbf{71.36} & \textbf{74.14}          & \textbf{98.32} & \textbf{95.31} & \textbf{96.79}          & \textbf{96.49} & \textbf{98.15} & \textbf{97.31}             \\
$\triangle$& \textcolor{blue}{(+2.21)} & \textcolor{blue}{(+2.32)} & \textcolor{blue}{(+2.28)} & \textcolor{blue}{(+0.26)} & \textcolor{blue}{(+0.88)} & \textcolor{blue}{(+0.60)} & \textcolor{blue}{(+0.41)} & \textcolor{blue}{(+0.92)} & \textcolor{blue}{(+0.67)} & \textcolor{blue}{(+1.05)} & \textcolor{blue}{(+0.13)} & \textcolor{blue}{(+0.60)}\\
\midrule
*PromptSRC   
& 84.14 & 75.65 & 79.67  
& \textbf{77.67} & 70.22 & 73.82   
& 98.03 & 93.83 & 95.88     
& 95.36 & {97.30} & 96.32  \\
\cellcolor{cyan!15}\textbf{+MQAdapter} 
& \cellcolor{cyan!15}\textbf{84.86} & \cellcolor{cyan!15}\textbf{76.99} & \cellcolor{cyan!15}\textbf{80.73}     
& {76.59} & \textbf{71.25} & \textbf{73.82}     
& \textbf{98.26} & \textbf{94.76} & \textbf{96.48}   
& \textbf{96.23} & \textbf{97.65} & \textbf{96.93}             \\ 
$\triangle$& \textcolor{blue}{(+0.73)} & \textcolor{blue}{(+1.34)} & \textcolor{blue}{(+1.07)} & \textcolor{blue}{(-1.08)} & \textcolor{blue}{(+1.03)} & \textcolor{blue}{(+0.07)} & \textcolor{blue}{(+0.23)} & \textcolor{blue}{(+0.93)} & \textcolor{blue}{(+0.59)} & \textcolor{blue}{(+0.87)} & \textcolor{blue}{(+0.35)} & \textcolor{blue}{(+0.61)} \\
\midrule
*MMRL++                  & 85.46 & 77.96 & 81.54                        & \textbf{77.92} & 71.22 & 74.42       & 99.10 & 94.32 & 96.65          & 95.06& 97.71 & 96.37             \\
\cellcolor{cyan!15}\textbf{+MQAdapter} & \cellcolor{cyan!15}\textbf{86.09} & \cellcolor{cyan!15}\textbf{78.93} & \cellcolor{cyan!15}\textbf{82.36}                        & {77.86} & \textbf{71.69} & \textbf{74.65}          & \textbf{99.35} & \textbf{95.20} & \textbf{97.23}          & \textbf{96.70} & \textbf{97.65} & \textbf{97.18}             \\
$\triangle$& \textcolor{blue}{(+0.63)} & \textcolor{blue}{(+0.96)} & \textcolor{blue}{(+0.82)} & \textcolor{blue}{(-0.06)} & \textcolor{blue}{(+0.47)} & \textcolor{blue}{(+0.23)} & \textcolor{blue}{(+0.25)} & \textcolor{blue}{(+0.88)} & \textcolor{blue}{(+0.58)} & \textcolor{blue}{(+1.64)} & \textcolor{blue}{(-0.06)} & \textcolor{blue}{(+0.81)}\\
\midrule\midrule
\multirow{2.4}{*}{Method} & \multicolumn{3}{c|}{StanfordCars}             & \multicolumn{3}{c|}{Flowers102} & \multicolumn{3}{c|}{Food101}    & \multicolumn{3}{c}{FGVCAircraft}  \\ 
\cmidrule(lr){2-4}\cmidrule(lr){5-7}\cmidrule(lr){8-10}\cmidrule(lr){11-13}
                        & Base  & New   & H                            & Base  & New   & H              & Base  & New   & H              & Base  & New   & H                 \\ 
\midrule
CLIP &63.37 &74.89 &68.65 &72.08 &77.80 &74.83 &90.10 &91.22 &90.66 &27.19 &36.29 &31.09\\
COMMA  &73.48 &74.91 &73.96 &94.86 &75.13 &83.88 &90.42 &92.74 &91.84 &36.47 &34.23 &35.84\\
MMA  &78.50 &73.10 &75.70 &97.77 &75.93 &85.48 &90.13 &91.30 &90.71 &40.57 &36.33 &38.33\\
TCP   &80.80 &74.13 &77.32 &97.73 &75.57 &85.23 &90.57 &91.37 &90.97 &41.97 &34.43 &37.83 \\
PromptSRC+DePT  &78.26 &74.73 &76.46 &97.44 &74.89 &84.69 &90.61 &91.63 &91.12 &41.18 &35.63 &38.20 \\
PromptSRC+DPC     &82.28 &74.98 &78.46 &97.44 &73.19 &83.59 &91.40 &91.58 &91.49 &46.74 &35.33 &40.24 \\
PromptSRC+ATP    &79.25 &74.95 &77.04 &97.82 &77.02 &86.18 &90.77 &91.78 &91.27 &42.47 &37.01 &39.55 \\
\midrule
*MaPLe                  
& 71.83 & {74.27} & 73.03  
& 95.28 & 73.55 & 83.02       
& 90.77 & 91.88 & 91.32        
& 36.16 & 34.39 & 35.25             \\
\textbf{+MQAdapter}                   
& \textbf{73.64} & \textbf{75.27} & \textbf{74.45}     
& \textbf{96.30} & \textbf{75.96} & \textbf{84.93}    
& \textbf{91.10} & \textbf{92.31} & \textbf{91.70}    
& \textbf{44.66} & \textbf{34.97} & \textbf{39.23}             \\ 
$\triangle$& \textcolor{blue}{(+1.81)} & \textcolor{blue}{(+1.00)} & \textcolor{blue}{(+1.42)}  & \textcolor{blue}{(+1.02)} & \textcolor{blue}{(+2.41)} & \textcolor{blue}{(+1.91)}      & \textcolor{blue}{(+0.33)} & \textcolor{blue}{(+0.43)} & \textcolor{blue}{(+0.38)} & \textcolor{blue}{(+8.50)} & \textcolor{blue}{(+0.58)} & \textcolor{blue}{(+3.98)}\\
\midrule
*PromptSRC               
& 78.27 & \textbf{75.27} & {76.78}   
& {97.87} & 76.63 & 85.96         
& 90.80 & 91.47 & 91.13       
& {43.10} & 37.20 & 39.93             \\
\textbf{+MQAdapter}                   
& \textbf{79.64} & 74.92 & \textbf{77.21}    
& \textbf{98.48} & \textbf{77.59} & \textbf{86.80}  
& \textbf{90.96}& \textbf{91.93}& \textbf{91.44}      
& \textbf{44.54} & \textbf{38.09} & \textbf{41.06}             \\ 
$\triangle$& \textcolor{blue}{(+1.37)} & \textcolor{blue}{(-0.35)} & \textcolor{blue}{(+0.47)} & \textcolor{blue}{(+0.61)} & \textcolor{blue}{(+0.96)} & \textcolor{blue}{(+0.84)} & \textcolor{blue}{(+0.16)} & \textcolor{blue}{(+0.46)} & \textcolor{blue}{(+0.31)} & \textcolor{blue}{(+1.44)} & \textcolor{blue}{(+0.89)} & \textcolor{blue}{(+1.13)}\\
\midrule
*MMRL++              
& {81.38} & 75.22 & {78.18}   
& 97.72 & 77.33 & 86.59       
& {90.41} & {91.69} & {91.05}     
& 47.18 & {38.33} & {42.30}             \\
\textbf{+MQAdapter}                    
& \textbf{82.08} & \textbf{75.39} & \textbf{78.59}           
& \textbf{98.67} & \textbf{78.01} & \textbf{87.13}       
& \textbf{90.79} & \textbf{91.97} & \textbf{91.38}          
& \textbf{48.32} & \textbf{38.75} & \textbf{43.01}            \\ 
$\triangle$& \textcolor{blue}{(+0.70)} & \textcolor{blue}{(+0.17)} & \textcolor{blue}{(+0.41)} & \textcolor{blue}{(+0.95)} & \textcolor{blue}{(+0.68)} & \textcolor{blue}{(+0.55)} & \textcolor{blue}{(+0.38)} & \textcolor{blue}{(+0.28)} & \textcolor{blue}{(+0.33)} & \textcolor{blue}{(+1.14)} & \textcolor{blue}{(+0.42)} & \textcolor{blue}{(+0.71)}\\
\midrule\midrule
\multirow{2.4}{*}{Method} & \multicolumn{3}{c|}{SUN397}                   & \multicolumn{3}{c|}{DTD}        & \multicolumn{3}{c|}{EuroSAT}    & \multicolumn{3}{c}{UCF101}        \\ 
\cmidrule(lr){2-4}\cmidrule(lr){5-7}\cmidrule(lr){8-10}\cmidrule(lr){11-13}
                        & Base  & New   & H                            & Base  & New   & H              & Base  & New   & H              & Base  & New   & H                 \\ 
\midrule
CLIP &69.36 &75.35 &72.23 &53.24 &59.90 &56.37 &56.48 &64.05 &60.03 &70.53 &77.50 &73.85\\
COMMA &80.94 &79.32 &80.86 &81.04 &58.62 &68.32 &93.56 &74.26 &83.42 &84.06 &80.56 &81.84\\
MMA &82.27 &78.57 &80.38 &83.20 &65.63 &73.38 &85.46 &82.34 &83.87 &86.23 &80.03 &82.20\\
TCP  &82.63 &78.20 &80.35 &82.77 &58.07 &68.25 &91.63 &74.73 &82.32 &87.13 &80.77 &83.83 \\
PromptSRC+DePT &82.60 &78.82 &80.67 &83.64 &59.18 &69.32 &94.46 &71.01 &81.07 &85.54 &77.29 &81.20 \\
PromptSRC+DPC  &83.63 &78.08 &80.76 &86.88 &54.31 &66.84 &96.25 &74.73 &84.13 &88.99 &78.13 &83.21 \\
PromptSRC+ATP     &82.73 &78.64 &80.63 &83.22 &62.68 &71.50 &92.29 &76.42 &83.61 &87.15 &79.23 &83.00 \\
\midrule
*MaPLe               & \textbf{81.03} & 78.21 & 79.59                        & 77.55 & 59.14 & 67.10          & 88.08 & 76.50 & 81.88          & 82.54 & 75.01 & 78.59             \\
\textbf{+MQAdapter}                    & 80.82 & \textbf{79.54} & \textbf{80.18}                        & \textbf{79.86} & \textbf{63.77} & \textbf{70.91}          & \textbf{95.14} & \textbf{84.10} & \textbf{89.28}          & \textbf{84.28} & \textbf{80.04} & \textbf{82.11}             \\ 
$\triangle$ & \textcolor{blue}{(-0.21)} & \textcolor{blue}{(+1.33)} & \textcolor{blue}{(+0.59)} & \textcolor{blue}{(+2.31)} & \textcolor{blue}{(+4.63)} & \textcolor{blue}{(+3.81)} & \textcolor{blue}{(+7.06)} & \textcolor{blue}{(+7.60)} & \textcolor{blue}{(+7.40)} & \textcolor{blue}{(+1.74)} & \textcolor{blue}{(+5.03)} & \textcolor{blue}{(+3.52)}\\
\midrule
*PromptSRC               
& 82.47 & 78.73 & 80.56      
& {82.90} & {62.67} & {71.38}    
& {92.83} & 70.06 & 79.85         
& 86.17 & {78.67} & 82.25          \\
\textbf{+MQAdapter}                   
& \textbf{82.97} & \textbf{78.83} & \textbf{80.85}          
& \textbf{83.91} & \textbf{63.89} & \textbf{72.54}        
& \textbf{93.88} & \textbf{78.64} & \textbf{85.59}      
& \textbf{88.00} & \textbf{79.29} & \textbf{83.42}             \\ 
$\triangle$ & \textcolor{blue}{(+0.5)} & \textcolor{blue}{(+0.1)} & \textcolor{blue}{(+0.29)} & \textcolor{blue}{(+1.01)} & \textcolor{blue}{(+1.22)} & \textcolor{blue}{(+1.16)} & \textcolor{blue}{(+1.05)} & \textcolor{blue}{(+8.58)} & \textcolor{blue}{(+5.73)} & \textcolor{blue}{(+1.83)} & \textcolor{blue}{(+0.62)} & \textcolor{blue}{(+1.17)}\\
\midrule
*MMRL++               
& {82.69} & 79.91 & 81.28 
& 85.30 & 65.82 & 74.30        
& 96.79 & 85.08 & 90.56       
& {86.50} & 80.58 & 83.44             \\
\textbf{+MQAdapter}                    
& \textbf{83.27} & \textbf{80.14} & \textbf{81.68}     
& \textbf{86.00} & \textbf{68.12} & \textbf{76.03}       
& \textbf{96.00} & \textbf{89.56} & \textbf{92.67}        
& \textbf{87.90} & \textbf{81.72} & \textbf{84.70}             \\
$\triangle$ & \textcolor{blue}{(+0.58)} & \textcolor{blue}{(+0.23)} & \textcolor{blue}{(+0.40)} & \textcolor{blue}{(+0.70)} & \textcolor{blue}{(+2.30)} & \textcolor{blue}{(+1.73)} & \textcolor{blue}{(-0.79)} & \textcolor{blue}{(+4.48)} & \textcolor{blue}{(+2.11)} & \textcolor{blue}{(+1.40)} & \textcolor{blue}{(+1.14)} & \textcolor{blue}{(+1.26)}\\
\bottomrule
\end{tabular}
\end{table*}

\subsection{Comparison Results}

\subsubsection{Base-to-New Generalization}
We integrate MQAdapter into three representative fine-tuning methods: MaPLe~\cite{khattak2023maple}, PromptSRC~\cite{khattak2023self}, and MMRL++~\cite{guo2026mmrl}. We further compare MQAdapter-enhanced methods with several state-of-the-art VLM adaptation approaches, including CLIP~\cite{radford2021learning}, COMMA~\cite{hu2024comma}, MMA~\cite{zhang2024multi}, and recent plug-and-play methods, PromptSRC+DePT~\cite{zhang2024dept}, TCP~\cite{yao2024tcp}, PromptSRC+DPC~\cite{li2025dpc}, and PromptSRC+ATP~\cite{li2025atprompt}. We report accuracy on both base classes (Base) and new classes (New), and use their harmonic mean (H) to evaluate the comprehensive performance between them in Tab.~\ref{base_new}. MQAdapter improves the harmonic mean of MaPLe, PromptSRC, and MMRL++ from 78.03\%, 79.67\%, and 81.54\% to 80.31\%, 80.73\%, and 82.36\%, achieving gains of 2.28\%, 1.07\%, and 0.82\%, respectively. For MaPLe, MQAdapter brings significant improvements on several challenging datasets, such as EuroSAT, DTD, UCF101, and FGVCAircraft. For PromptSRC and MMRL++, MQAdapter also achieves consistent gains on all datasets, demonstrating its robustness across different adaptation frameworks. Moreover, MQAdapter also achieves the best average harmonic mean than existing plug-and-play approaches. These results demonstrate that MQAdapter effectively retrieves the Top-$K$ textual categories most relevant to the input image and uses them as semantic anchors to guide visual feature adaptation.  By further exploiting the cross-modal interactions between the visual representation and these semantic anchors via quantum computation, MQAdapter learns more discriminative visual features and improves generalization on both base and new classes.

\subsubsection{Few-shot Image Classification}
We further evaluate MQAdapter on few-shot image classification under 1-, 2-, 4-, 8-, and 16-shot settings. MQAdapter is integrated into three representative frameworks, i.e., MaPLe~\cite{khattak2023maple}, PromptSRC~\cite{khattak2023self}, and MMRL++~\cite{guo2026mmrl}. The results are reported in Tab.~\ref{tab_few_shot}. We compare our methods with recent competitive few-shot adaptation methods, including CoOp~\cite{zhou2022learning}, CoCoOp~\cite{zhou2022conditional}, GraphAdapter~\cite{li2024graphadapter}, MaPLe~\cite{khattak2023maple}, PromptSRC~\cite{khattak2023self}, and MMRL++~\cite{guo2026mmrl}. As shown in Tab.~\ref{tab_few_shot}, MQAdapter consistently improves the performance of MaPLe, PromptSRC, and MMRL++ across different shot settings. Specifically, when combined with MaPLe, MQAdapter improves the average accuracy by 3.75\%, 2.91\%, 2.07\%, 1.34\%, and 0.94\% under the 1-, 2-, 4-, 8-, and 16-shot settings, respectively. For PromptSRC, the corresponding gains are 1.94\%, 1.74\%, 0.96\%, 0.89\%, and 0.72\%. For MMRL++, MQAdapter brings improvements of 1.49\%, 1.60\%, 0.79\%, 1.05\%, and 0.76\%, respectively. MQAdapter achieves the largest gains in low-shot scenarios. For example, MQAdapter improves MaPLe, PromptSRC, and MMRL++ by 3.75\%, 1.94\%, and 1.49\% under the 1-shot setting, and by 2.91\%, 1.74\%, and 1.60\% under the 2-shot setting, respectively. The consistent gains across different baselines, datasets, and shot numbers verify the robustness and effectiveness of MQAdapter for few-shot VLM adaptation.

\begin{table*}[htbp]
  \centering
  \caption{Numerical comparison results for few-shot learning across 1-/2-/4-/8-/16-shot settings on 11 datasets.}
   \resizebox{1\textwidth}{!}{
    \begin{tabular}{c|c|ccccccccccc|c}
    \toprule
    \rotatebox{90}{Method} & \rotatebox{90}{Shot} & \rotatebox{90}{Caltech101} & \rotatebox{90}{DTD}   & \rotatebox{90}{EuroSAT} & \rotatebox{90}{FGVCAircraft} & \rotatebox{90}{Flowers102} & \rotatebox{90}{Food101} & \rotatebox{90}{ImageNet} &  \rotatebox{90}{OxfordPets} & \rotatebox{90}{StanfordCars} & \rotatebox{90}{SUN397} &  \rotatebox{90}{UCF101} & \rotatebox{90}{Average} \\
    \midrule
    {CoOp } & \multirow{12}[1]{*}{1} & 92.60  & 50.23 & 54.93 & 21.37 & 77.53 & 84.33 & 66.33 & 90.37 & 67.43 & 66.77 & 71.23 & 67.56  \\
    {CoCoOp} &       & 93.83 & 48.54 & 55.33 & 12.68 & 72.08 & 85.65 & 69.43 & 91.27 & 67.22 & 68.33 & 70.30  & 66.79  \\
    {GraphAdapter} & &92.94	&52.72	&67.54	&29.37	&86.93	&85.24	&69.01	&88.17	&66.15	&66.50	&72.40 &70.63\\
    {MaPLe} &       & 92.57 & 52.13 & 71.80  & 26.73 & 83.30  & 80.50  & 62.67 & 89.10  & 66.60  & 64.77 & 71.83 & 69.27  \\
    % \rowcolor{gray!20}
    \rowcolor{gray!20} {+MQAdapter } & & 95.01 & 54.37 & 77.72 & 30.30 & 84.86 & 85.25 & 70.01 & 92.50 & 69.08 & 68.91 & 75.20 & 73.02 \\
         $\triangle$ & & \textcolor{blue}{(+2.44)} &\textcolor{blue}{(+2.24)} &\textcolor{blue}{(+5.92)} &\textcolor{blue}{(+3.57)} &\textcolor{blue}{(+1.56)} &\textcolor{blue}{(+4.75)} &\textcolor{blue}{(+7.34)} &\textcolor{blue}{(+3.40)} &\textcolor{blue}{(+2.48)} &\textcolor{blue}{(+4.14)} &\textcolor{blue}{(+3.37)} &\textcolor{blue}{(+3.75)} \\
    {PromptSRC} &       & 93.67 & 56.23 & 73.13 & 27.67 & 85.93 & 84.87 & 68.13 & 92.00  & 69.40  & 69.67 & 74.80  & 72.32  \\
    \rowcolor{gray!20}
{+MQAdapter } &       & 95.24 & 58.69 & 78.93 & 30.78 & 87.01 & 86.57 & 70.31 & 92.97 & 70.99 & 69.72 & 75.60 & 74.26  \\
$\triangle$ &       &\textcolor{blue}{(+1.57)} &\textcolor{blue}{(+2.46)} &\textcolor{blue}{(+5.80)} &\textcolor{blue}{(+3.11)} &\textcolor{blue}{(+1.08)} &\textcolor{blue}{(+1.70)} &\textcolor{blue}{(+2.18)} &\textcolor{blue}{(+0.97)} &\textcolor{blue}{(+1.59)} &\textcolor{blue}{(+0.05)} &\textcolor{blue}{(+0.80)} &\textcolor{blue}{(+1.94)} \\
    {MMRL++} & & 94.07 & 57.00 & 79.07 & 28.10 & 84.13 & 83.23 & 70.03 & 90.97 & 68.20 & 68.97 & 75.50 & 72.66 \\
   \rowcolor{gray!20} {+MQAdapter } & & 95.29 & 59.69 & 82.32 & 30.09 & 85.55 & 84.38 & 70.59 & 91.36 & 69.44 & 69.70 & 77.29 & 74.15 \\
$\triangle$ & & \textcolor{blue}{(+1.22)} &\textcolor{blue}{(+2.69)} &\textcolor{blue}{(+3.25)} &\textcolor{blue}{(+1.99)} &\textcolor{blue}{(+1.42)} &\textcolor{blue}{(+1.15)} &\textcolor{blue}{(+0.56)} &\textcolor{blue}{(+0.39)} &\textcolor{blue}{(+1.24)} &\textcolor{blue}{(+0.73)} &\textcolor{blue}{(+1.79)} &\textcolor{blue}{(+1.49)} \\
    \midrule
    {CoOp } & \multirow{12}[2]{*}{2} & 93.07 & 53.60  & 65.17 & 26.20  & 87.33 & 84.40  & 67.07 & 89.80  & 70.50  & 66.53 & 73.43 & 70.65   \\
    {CoCoOp} &       & 94.82 & 52.17 & 46.74 & 15.06 & 75.79 & 86.22 & 69.78 & 92.64 & 68.37 & 69.03 & 73.51 & 67.65  \\
    {GraphAdapter} & & 94.36	&57.39	&71.74	&32.70	&91.07	&85.84	&69.48	&91.14	&71.45	&69.45	&76.55 &73.74\\
    {MaPLe} &       & 93.97 & 55.50  & 78.30  & 30.90  & 88.93 & 81.47 & 65.10  & 90.87 & 71.60  & 67.10  & 74.60  & 72.58  \\
    % \rowcolor{gray!20}
     \rowcolor{gray!20} {+MQAdapter } & & 95.29 & 59.93 & 82.46 & 33.06 & 89.20 & 86.50 & 70.45 & 93.38 & 71.98 & 70.80 & 77.29 & 75.49 \\
$\triangle$ & & \textcolor{blue}{(+1.32)} &\textcolor{blue}{(+4.43)} &\textcolor{blue}{(+4.16)} &\textcolor{blue}{(+2.16)} &\textcolor{blue}{(+0.27)} &\textcolor{blue}{(+5.03)} &\textcolor{blue}{(+5.35)} &\textcolor{blue}{(+2.51)} &\textcolor{blue}{(+0.38)} &\textcolor{blue}{(+3.70)} &\textcolor{blue}{(+2.69)} &\textcolor{blue}{(+2.91)} \\

    {PromptSRC} &       & 94.53 & 59.97 & 79.37 & 31.70  & 91.17 & 85.70  & 69.77 & 92.50  & 73.40  & 71.60  & 78.50  & 75.29  \\
    % \rowcolor{gray!20}
    \rowcolor{gray!20}
{+MQAdapter } &       & 95.54 & 62.41 & 85.78 & 34.17 & 92.61 & 87.12 & 70.52 & 93.19 & 74.06 & 72.02 & 79.96 & 77.03  \\
$\triangle$ &       &\textcolor{blue}{(+1.01)} &\textcolor{blue}{(+2.44)} &\textcolor{blue}{(+6.41)} &\textcolor{blue}{(+2.47)} &\textcolor{blue}{(+1.44)} &\textcolor{blue}{(+1.42)} &\textcolor{blue}{(+0.75)} &\textcolor{blue}{(+0.69)} &\textcolor{blue}{(+0.66)} &\textcolor{blue}{(+0.42)} &\textcolor{blue}{(+1.46)} &\textcolor{blue}{(+1.74)} \\
    {MMRL++} & & 94.90 & 60.80 & 81.17 & 32.40 & 89.70 & 84.07 & 70.97 & 91.17 & 72.70 & 70.97 & 78.87 & 75.25 \\
 \rowcolor{gray!20} {+MQAdapter } & & 95.86 & 63.42 & 86.80 & 34.53 & 90.99 & 84.84 & 71.02 & 92.29 & 73.75 & 71.80 & 80.02 & 76.85 \\
$\triangle$ & & \textcolor{blue}{(+0.96)} &\textcolor{blue}{(+2.62)} &\textcolor{blue}{(+5.63)} &\textcolor{blue}{(+2.13)} &\textcolor{blue}{(+1.29)} &\textcolor{blue}{(+0.77)} &\textcolor{blue}{(+0.05)} &\textcolor{blue}{(+1.12)} &\textcolor{blue}{(+1.05)} &\textcolor{blue}{(+0.83)} &\textcolor{blue}{(+1.15)} &\textcolor{blue}{(+1.60)} \\

    \midrule
    {CoOp } & \multirow{12}[2]{*}{4} & 94.40  & 58.70  & 70.80  & 30.83 & 92.17 & 84.47 & 68.73 & 92.57 & 74.47 & 69.97 & 77.10  & 74.02   \\
    {CoCoOp} &       & 94.98 & 55.04 & 65.56 & 24.79 & 78.40  & 86.88 & 70.39 & 92.81 & 69.39 & 70.21 & 74.82 & 71.21  \\
    {GraphAdapter} & & 95.01	&62.83	&78.20	&35.88	&95.01	&86.32	&70.58	&91.99	&74.07	&72.36	&80.89 &76.65\\
    {MaPLe} &       & 94.43 & 61.00  & 84.50  & 34.87 & 92.67 & 81.77 & 67.70  & 91.90  & 75.30  & 70.67 & 78.47 & 75.75  \\
    % \rowcolor{gray!20}
    \rowcolor{gray!20} {+MQAdapter } & & 95.38 & 63.71 & 86.91 & 36.48 & 93.50 & 86.81 & 70.93 & 93.51 & 75.68 & 72.98 & 80.17 & 77.82 \\
$\triangle$ & & \textcolor{blue}{(+0.95)} &\textcolor{blue}{(+2.71)} &\textcolor{blue}{(+2.41)} &\textcolor{blue}{(+1.61)} &\textcolor{blue}{(+0.83)} &\textcolor{blue}{(+5.04)} &\textcolor{blue}{(+3.23)} &\textcolor{blue}{(+1.61)} &\textcolor{blue}{(+0.38)} &\textcolor{blue}{(+2.31)} &\textcolor{blue}{(+1.70)} &\textcolor{blue}{(+2.07)} \\

    {PromptSRC} &       & 95.27 & 65.53 & 86.30  & 37.47 & 93.87 & 86.17 & 71.07 & 93.43 & 77.13 & 74.00  & 81.57 & 78.35  \\
    % \rowcolor{gray!20}
    \rowcolor{gray!20}
{+MQAdapter } &       & 96.06 & 66.19 & 88.89 & 39.03 & 95.57 & 87.20 & 71.13 & 93.73 & 77.75 & 74.43 & 82.34 & 79.30  \\
$\triangle$ &       &\textcolor{blue}{(+0.79)} &\textcolor{blue}{(+0.66)} &\textcolor{blue}{(+2.59)} &\textcolor{blue}{(+1.56)} &\textcolor{blue}{(+1.70)} &\textcolor{blue}{(+1.03)} &\textcolor{blue}{(+0.06)} &\textcolor{blue}{(+0.30)} &\textcolor{blue}{(+0.62)} &\textcolor{blue}{(+0.43)} &\textcolor{blue}{(+0.77)} &\textcolor{blue}{(+0.96)} \\
   {MMRL++} & & 95.80 & 67.03 & 89.23 & 40.87 & 93.60 & 84.83 & 71.47 & 92.43 & 77.83 & 73.43 & 82.50 & 79.00 \\
 \rowcolor{gray!20} {+MQAdapter } & & 96.23 & 68.32 & 90.10 & 41.82 & 94.52 & 85.68 & 71.83 & 93.08 & 78.58 & 74.02 & 83.48 & 79.79 \\
$\triangle$ & & \textcolor{blue}{(+0.43)} &\textcolor{blue}{(+1.29)} &\textcolor{blue}{(+0.87)} &\textcolor{blue}{(+0.95)} &\textcolor{blue}{(+0.92)} &\textcolor{blue}{(+0.85)} &\textcolor{blue}{(+0.36)} &\textcolor{blue}{(+0.65)} &\textcolor{blue}{(+0.75)} &\textcolor{blue}{(+0.59)} &\textcolor{blue}{(+0.98)} &\textcolor{blue}{(+0.79)} \\

    \midrule
    {CoOp } & \multirow{12}[2]{*}{8} & 94.37 & 64.77 & 78.07 & 39.00  & 94.97 & 82.67 & 70.63 & 91.27 & 79.30  & 71.53 & 80.20  & 76.98   \\
    {CoCoOp} &       & 95.04 & 58.89 & 68.21 & 26.61 & 84.30  & 86.97 & 70.63 & 93.45 & 70.44 & 70.84 & 77.14 & 72.96  \\
    {GraphAdapter} & & 95.29	&67.79	&80.53	&40.02	&96.47	&87.03	&71.71	&92.15	&78.06	&74.14	&82.53 &78.70\\
    {MaPLe} &       & 95.20  & 66.50  & 87.73 & 42.00  & 95.80  & 83.60  & 70.30  & 92.57 & 79.47 & 73.23 & 81.37 & 78.89  \\
    % \rowcolor{gray!20}
     \rowcolor{gray!20} {+MQAdapter } & & 95.98 & 67.67 & 90.09 & 42.46 & 96.51 & 87.06 & 71.31 & 94.00 & 79.83 & 74.47 & 83.19 & 80.23 \\
$\triangle$ & & \textcolor{blue}{(+0.78)} &\textcolor{blue}{(+1.17)} &\textcolor{blue}{(+2.36)} &\textcolor{blue}{(+0.46)} &\textcolor{blue}{(+0.71)} &\textcolor{blue}{(+3.46)} &\textcolor{blue}{(+1.01)} &\textcolor{blue}{(+1.43)} &\textcolor{blue}{(+0.36)} &\textcolor{blue}{(+1.24)} &\textcolor{blue}{(+1.82)} &\textcolor{blue}{(+1.34)} \\

    {PromptSRC} &       & 95.67 & 69.87 & 88.80  & 43.27 & 96.27 & 86.90  & 72.33 & 93.50  & 80.97 & 75.73  & 84.30  & 80.69  \\
    % \rowcolor{gray!20}
    \rowcolor{gray!20}
{+MQAdapter } &       & 96.11 & 70.51 & 92.30 & 45.06 & 96.59 & 87.50 & 72.82 & 94.39 & 81.53 & 75.92 & 84.77 & 81.59  \\
$\triangle$ &       &\textcolor{blue}{(+0.44)} &\textcolor{blue}{(+0.64)} &\textcolor{blue}{(+3.50)} &\textcolor{blue}{(+1.79)} &\textcolor{blue}{(+0.32)} &\textcolor{blue}{(+0.60)} &\textcolor{blue}{(+0.49)} &\textcolor{blue}{(+0.79)} &\textcolor{blue}{(+0.56)} &\textcolor{blue}{(+0.19)} &\textcolor{blue}{(+0.47)} &\textcolor{blue}{(+0.89)} \\
    {MMRL++} & & 96.13 & 70.83 & 89.17 & 49.23 & 96.33 & 85.57 & 72.17 & 92.70 & 82.40 & 75.53 & 84.53 & 81.33 \\
\rowcolor{gray!20} {+MQAdapter } & & 97.12 & 71.63 & 91.93 & 51.01 & 97.12 & 86.07 & 72.49 & 93.62 & 83.59 & 75.74 & 85.88 & 82.38 \\
$\triangle$ & & \textcolor{blue}{(+0.99)} &\textcolor{blue}{(+0.80)} &\textcolor{blue}{(+2.76)} &\textcolor{blue}{(+1.78)} &\textcolor{blue}{(+0.79)} &\textcolor{blue}{(+0.50)} &\textcolor{blue}{(+0.32)} &\textcolor{blue}{(+0.92)} &\textcolor{blue}{(+1.19)} &\textcolor{blue}{(+0.21)} &\textcolor{blue}{(+1.35)} &\textcolor{blue}{(+1.05)} \\

    \midrule
    {CoOp } & \multirow{12}[2]{*}{16} & 95.57 & 69.87 & 84.93 & 43.40  & 97.07 & 84.20  & 71.87 & 91.87 & 83.07 & 74.67 & 82.23 & 79.89   \\
    {CoCoOp} &       & 95.16 & 63.04 & 73.32 & 31.21 & 87.84 & 87.25 & 70.83 & 93.34 & 71.57 & 72.15 & 78.14 & 74.90  \\
    {GraphAdapter} & & 95.66	&72.40	&86.21	&44.97	&97.77	&87.27	&73.40	&92.83	&83.85	&75.19	&84.54 &81.28\\
    {MaPLe} &       & 96.00  & 71.33 & 92.33 & 48.40  & 97.00  & 85.33 & 72.33 & 92.83 & 83.57 & 75.53 & 85.03 & 81.79  \\
    % \rowcolor{gray!20}
     \rowcolor{gray!20} {+MQAdapter } & & 96.55 & 73.29 & 93.40 & 48.81 & 97.32 & 87.83 & 72.67 & 94.63 & 83.79 & 75.78 & 85.93 & 82.73 \\
$\triangle$ & & \textcolor{blue}{(+0.55)} &\textcolor{blue}{(+1.96)} &\textcolor{blue}{(+1.07)} &\textcolor{blue}{(+0.41)} &\textcolor{blue}{(+0.32)} &\textcolor{blue}{(+2.50)} &\textcolor{blue}{(+0.34)} &\textcolor{blue}{(+1.80)} &\textcolor{blue}{(+0.22)} &\textcolor{blue}{(+0.25)} &\textcolor{blue}{(+0.90)} &\textcolor{blue}{(+0.94)} \\

    {PromptSRC} &       & 96.07 & 72.73 & 92.43 & 50.83 & 97.60  & 87.50  & 73.17 & 93.67 & 83.83 & 77.23 & 86.47 & 82.87  \\
    % \rowcolor{gray!20}
    \rowcolor{gray!20}
{+MQAdapter} &       & 96.67 & 74.29 & 94.77 & 51.25 & 97.97 & 87.82 & 73.24 & 94.58 & 84.50 & 77.45 & 86.88 & 83.58  \\
$\triangle$ &       &\textcolor{blue}{(+0.60)} &\textcolor{blue}{(+1.56)} &\textcolor{blue}{(+2.34)} &\textcolor{blue}{(+0.42)} &\textcolor{blue}{(+0.37)} &\textcolor{blue}{(+0.32)} &\textcolor{blue}{(+0.07)} &\textcolor{blue}{(+0.91)} &\textcolor{blue}{(+0.67)} &\textcolor{blue}{(+0.22)} &\textcolor{blue}{(+0.41)} &\textcolor{blue}{(+0.72)} \\
    {MMRL++} & & 96.73 & 74.37 & 93.50 & 58.20 & 98.20 & 86.07 & 73.00 & 93.30 & 86.20 & 77.50 & 87.43 & 84.05 \\
 \rowcolor{gray!20} {+MQAdapter } & & 97.57 & 76.71 & 94.17 & 59.35 & 98.50 & 86.54 & 73.30 & 94.44 & 86.71 & 77.82 & 87.79 & 84.81 \\
$\triangle$ & & \textcolor{blue}{(+0.84)} &\textcolor{blue}{(+2.34)} &\textcolor{blue}{(+0.67)} &\textcolor{blue}{(+1.15)} &\textcolor{blue}{(+0.30)} &\textcolor{blue}{(+0.47)} &\textcolor{blue}{(+0.30)} &\textcolor{blue}{(+1.14)} &\textcolor{blue}{(+0.51)} &\textcolor{blue}{(+0.32)} &\textcolor{blue}{(+0.36)} &\textcolor{blue}{(+0.76)} \\     
    \bottomrule
    \end{tabular}%
  \label{tab_few_shot}%
  }
\end{table*}%

\begin{table*}[!h]
\caption{Cross-dataset transfer experiments of three baselines w/ or w/o our MQAdapter on 11 datasets.}
\centering
\renewcommand{\arraystretch}{0.8}
\setlength\tabcolsep{2pt}
\begin{tabular}{c|c|c|cccccccccc} 
\toprule
\multirow{2}{*}{Method} & \textbf{Source} & \multicolumn{11}{c}{\textbf{Target}}\\ 
\cmidrule(lr){2-2}\cmidrule(lr){3-13}
& ImageNet & Avg & Caltech101 & OxfordPets & StanfordCars & Flowers102 & Food101 & FGVCAircraft & SUN397 & DTD & EuroSAT & UCF101 \\ 
\midrule
CoCoOp  & 71.02 & 65.74 & 94.43 & 90.14 & 65.32 & 71.88 & 86.06 & 22.94 & 67.36 & 45.73 & 45.37 & 68.21 \\ 
MMA     & 71.00 & 66.61 & 93.80 & 90.30 & 66.13 & 72.07 & 86.12 & 25.33 & 68.17 & 46.57 & 49.24 & 68.32 \\
TCP     & 71.40 & 66.29 & 93.97 & 91.25 & 64.69 & 71.21 & 86.69 & 23.45 & 67.15 & 44.35 & 51.45 & 68.73 \\
\midrule
MaPLe   & 70.72 & 66.30 & 93.53 & 90.49 & 65.57 & 72.23 & 86.20 & \textbf{24.74} & 67.01 & \textbf{46.49} & 48.06 & 68.69 \\
\rowcolor{gray!20}
\textbf{+MQAdapter} & 70.81 & \textbf{66.92} & \textbf{94.32} & \textbf{90.95} & \textbf{65.60} & \textbf{72.39} & \textbf{86.30} & 24.57 & \textbf{67.11} & 46.45 & \textbf{52.28} & \textbf{69.18} \\
\midrule
PromptSRC & 71.27 & 65.81 & 93.60 & 90.25 & 65.70 & 70.25 & 86.15 & 23.90 & 67.10 & 46.87 & 45.50 & 68.75 \\
\rowcolor{gray!20}
\textbf{+MQAdapter} & 70.75 & \textbf{66.26} & \textbf{93.96} & \textbf{90.65} & \textbf{66.05} & \textbf{71.42} & \textbf{86.45} & \textbf{24.90} & \textbf{67.31} & \textbf{47.28} & 45.07 & \textbf{69.52} \\
\midrule
MMRL++ & 71.87 & 67.49 & 94.63 & 91.43 & 66.60 & 73.53 & 86.73 & 26.07 & 67.77 & 46.13 & 53.00 & 69.03 \\
\rowcolor{gray!20}
\textbf{+MQAdapter} & 72.07 & \textbf{68.10} & \textbf{94.93} & \textbf{91.61} & \textbf{66.73} & \textbf{73.77} & \textbf{86.80} & \textbf{26.76} & \textbf{68.19} & \textbf{46.22} & \textbf{56.26} & \textbf{69.76} \\
\bottomrule
\end{tabular}
\label{cross}
\end{table*}
\subsubsection{Cross-Dataset Transfer and Domain Generalization}
As shown in Tab.~\ref{cross}, MQAdapter consistently improves cross-dataset transfer performance when applied to MaPLe, PromptSRC, and MMRL++. The average accuracy increases by 0.62\%, 0.45\%, and 0.61\%, respectively. Tab.~\ref{tab2} further shows that MQAdapter also improves domain generalization for all baselines across different ImageNet variants. 
These results indicate that MQAdapter enhances adaptation performance while preserving the transferability of pre-trained VLMs.

\begin{table}[!htbp]
\caption{Cross-domain generalization experiments of three baselines w/ or w/o MQAdapter.}
\centering
\renewcommand{\arraystretch}{0.8}
\begin{tabular}{c|c|cccc} 
\toprule
\multirow{2}{*}{Method} & \textbf{Source} & \multicolumn{4}{c}{\textbf{Target}}\\ 
\cmidrule(lr){2-2}\cmidrule(lr){3-6}
& ImageNet  & -V2 & -Sketch & -A & -R \\
\midrule
CoCoOp & 71.02 & 64.07 & 48.75 & 50.63 & 76.18 \\
MMA & 71.00 & 64.33 & 49.13 & 51.12 & 77.32 \\
\midrule
MaPLe & 70.72 & 64.07 & 49.15 & 50.90 & 76.98 \\
\rowcolor{gray!20}
\textbf{+MQAdapter} & 70.56 & \textbf{64.24} & \textbf{49.33} & \textbf{51.41} & \textbf{77.21} \\
\midrule
PromptSRC & 71.27 & 64.35 & 49.55 & 50.90 & {77.80} \\
\rowcolor{gray!20}
\textbf{+MQAdapter} & 70.81 & \textbf{64.53} & \textbf{49.65} & \textbf{51.15} & \textbf{77.90} \\
\midrule
MMRL++ & 71.87 & 64.67 & 49.30 & 51.00 & 77.43 \\
\rowcolor{gray!20}
\textbf{+MQAdapter} & 71.99 & \textbf{64.77} & \textbf{49.56} & \textbf{51.64} & \textbf{77.55} \\
\bottomrule
\end{tabular}
\label{tab2}
\end{table}

% ----------------------

\begin{table}[!h]
\caption{Ablation study of MQAdapter in the base-to-new task based on MMRL++. ``cross-attention'' denotes conventional cross-modal interaction in the classical feature space. }
\label{CQL}
    \centering
    \renewcommand{\arraystretch}{0.8}
\begin{tabular}{c|ccc} 
\toprule
\multirow{2.4}{*}{Method} & \multicolumn{3}{c}{\textbf{Avg. over 11 datasets}}   \\ 
\cmidrule(lr){2-4}
                & Base  & New   & H                          \\ 
\midrule
Baseline              & 85.46 & 77.96 & 81.54      \\
\midrule
Baseline+cross-attention         & 85.89 & 78.03 & 81.77   \\
Baseline+separate VQC        & 85.54 & 78.47 & 81.85   \\
Baseline+MQAdapter      & 86.09 & 78.93 & 82.36   \\
\bottomrule
\end{tabular}
\end{table}

\subsection{Ablation Experiments}
\subsubsection{Effectiveness of Cross-modal Quantum Learning mechanism}
To evaluate the effectiveness of the cross-modal quantum learning mechanism, we conduct ablation experiments on the MMRL++ baseline. Specifically, we replace the cross-modal quantum learning mechanism with a conventional cross-attention module~\cite{vaswani2017attention} to perform fine-stage cross-modal interaction. This variant improves the harmonic mean by 0.23\%, indicating that the text semantic anchors can provide useful guidance for visual feature learning. The proposed MQAdapter achieves the harmonic mean of 82.36\%, outperforming the baseline by 0.82\% and conventional cross-attention by 0.59\%. This demonstrates that MQAdapter learns more discriminative visual representations than conventional cross-attention by effectively modeling higher-order cross-modal interactions. Our  MQAdapter use a shared VQC to align different modalities and facilitate cross-modal interaction in a unified quantum representation space. To validate the effectiveness of this design, we conduct an ablation study by replacing the shared VQC with separate VQCs. The results show that the shared VQC achieves better performance than the separate VQC, demonstrating the effectiveness of our cross-modal quantum learning mechanism while requiring fewer quantum resources.

% By modeling text-visual interactions in the quantum feature space, CQA captures higher-order cross-modal correlations and provides more effective semantic guidance for visual feature refinement. As a result, it learns more discriminative visual representations and further improves performance.

\subsubsection{Influence of hyper-parameters in MQAdapter}
\begin{figure}[htbp]
  \centering
  \includegraphics[width=1\linewidth]{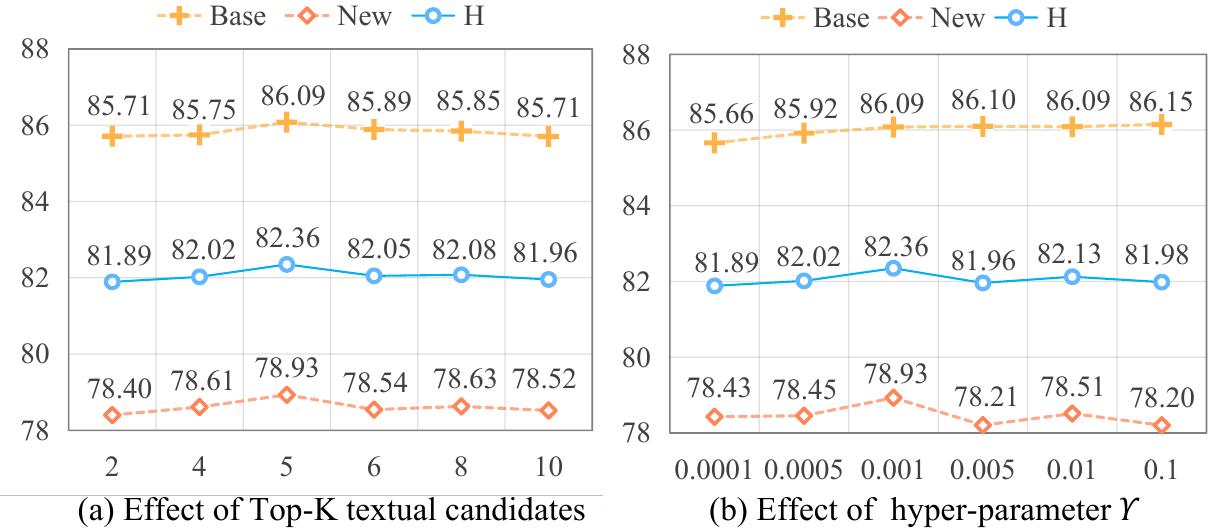}
\caption{ Hyper-parameter analysis of MQAdapter based on the MMRL++ baseline, averaged over 11 datasets for base-to-new generalization.} 
\label{hyper}
\end{figure}

To evaluate the influence of hyper-parameters in MQAdapter, we perform experiments on the MMRL++ baseline. Fig.~\ref{hyper} (a) presents the average base-to-new generalization performance on 11 datasets under different choices of Top-$K$ textual candidates. When $K$ increases from 2 to 5, the harmonic mean improves from 81.89\% to 82.36\%. This indicates that semantically relevant textual candidates provide effective guidance for visual feature refinement. However, further increasing $K$ does not bring additional gains. This suggests that more candidates may introduce less relevant categories and weaken the discriminative guidance for visual representation learning. Therefore, we set $K=5$ in the final model.
Fig.~\ref{hyper} (b) shows the influence of the scaling factor $\gamma$. MQAdapter achieves the best harmonic mean of 82.36\% when $\gamma=0.001$. A smaller $\gamma$ may limit the model's ability to adapt to downstream tasks. In contrast, a larger $\gamma$ increases the reliance on task-specific features, which may reduce transferability to new classes.

\begin{figure}[htbp]
  \centering
  \includegraphics[width=0.8\linewidth]{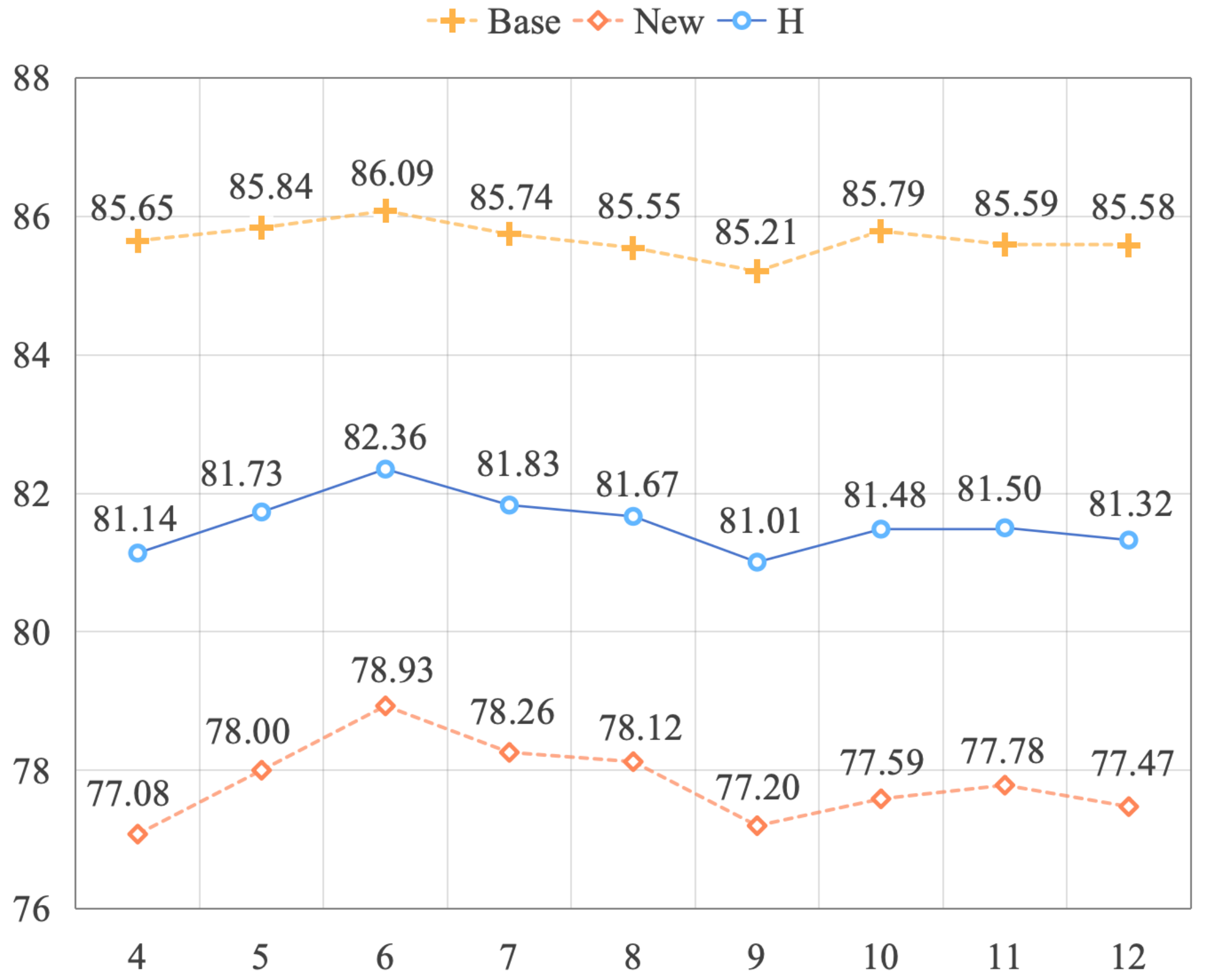}
\caption{Effect of the MQAdapter insertion start depth in the CLIP visual encoder. Results are obtained using MMRL++ as the baseline and averaged over 11 datasets for base-to-new generalization. } 
\label{layer}
\end{figure}

\subsubsection{Influence of MQAdapter insertion depth}
We further analyze the impact of inserting MQAdapter at different layers of the visual encoder. The experiments are conducted on the MMRL++ baseline. We report average base-to-new generalization performance
on 11 datasets. As shown in Fig.~\ref{layer}, we obtain the best performance when inserted from layer 6. When MQAdapter is inserted from shallower layers, the accuracy decreases on both base and new classes. This suggests that early CLIP layers mainly capture general visual information and are less suitable for task-specific adaptation. In contrast, when MQAdapter is inserted from deeper layers, only a few blocks can be adapted, which may be insufficient to progressively enhance visual representations.
These results show that inserting MQAdapter at intermediate layers provides a better balance between preserving CLIP's general knowledge and adapting to downstream tasks.

\begin{figure*}[htbp]
  \centering
  \includegraphics[width=0.8\linewidth]{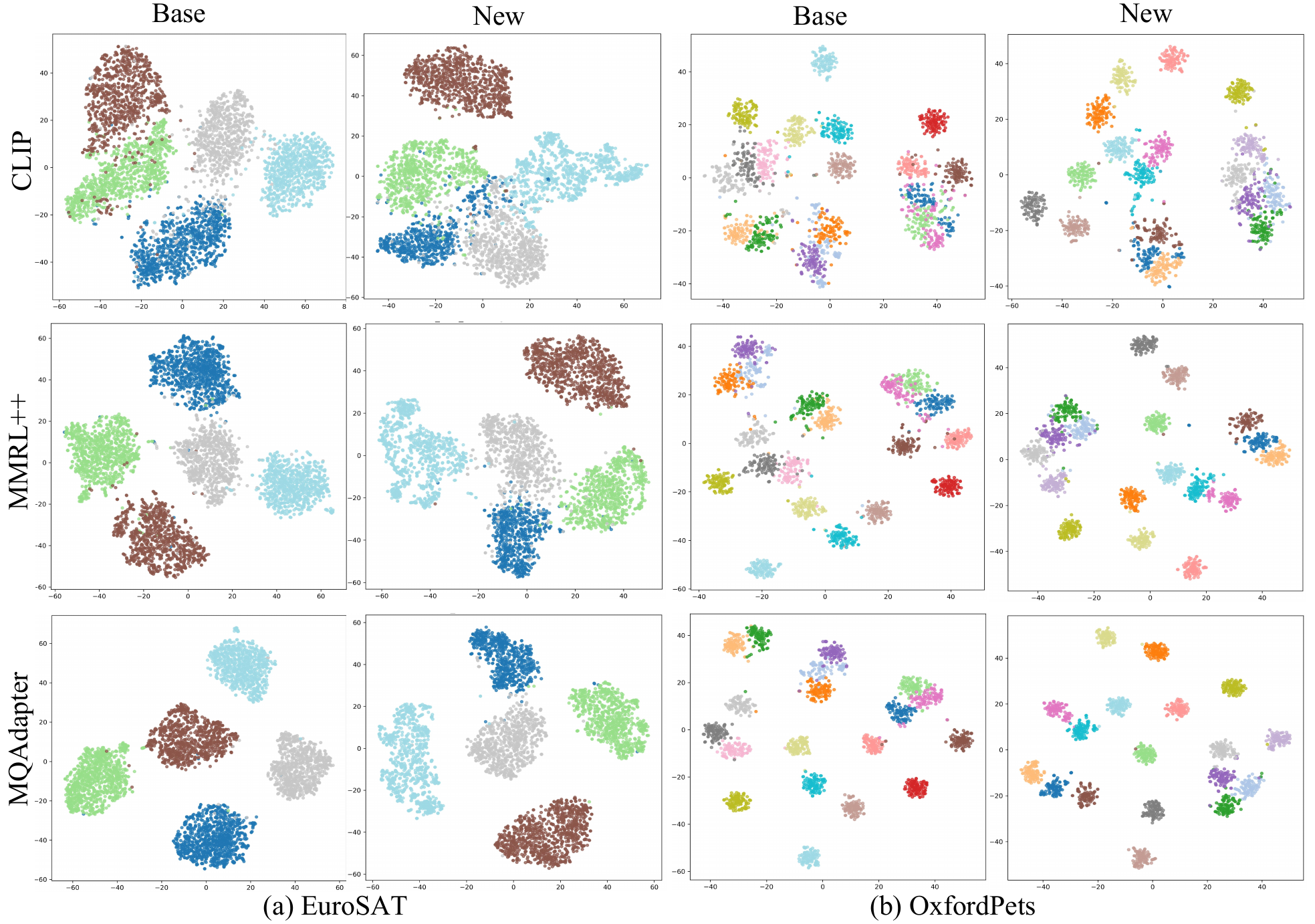}
\caption{t-SNE visualizations of learned visual features on the base and new classes of the EuroSAT and OxfordPets datasets. Each color denotes one class. Our MQAdapter improves intra-class compactness and inter-class separability.} 
\label{visual}
\end{figure*}

\subsection{Training Stability}
To assess the training stability of our MQAdapter, we integrated it into the MMRL++ baseline and report the mean performance and standard deviation (std) over three random seeds. As shown in Table~\ref{tab:imagenet_eurosat}, +MQAdapter achieves lower standard deviation than MMRL++ while maintaining superior performance across the evaluated datasets. This indicates that MQAdapter not only improves adaptation accuracy but also enables a more stable optimization process under different random initializations.

\begin{table}[htbp]
\centering
\caption{Training stability comparison across different datasets.}
\label{tab:imagenet_eurosat}
\resizebox{\linewidth}{!}{
\begin{tabular}{l|cc|cc}
\toprule
\multirow{2}{*}{Method} 
& \multicolumn{2}{c|}{ImageNet} 
& \multicolumn{2}{c}{Caltech101} \\
& Base & New 
& Base & New  \\
\midrule
MMRL++ 
& $77.92 \pm {0.12}$ 
& $71.22 \pm 0.05$ 
& $99.10 \pm 0.08$ 
& $94.32 \pm 0.22$ \\

\textbf{+MQAdapter}
& $77.86 \pm 0.06$ 
& $71.69 \pm {0.09}$ 
& $99.35 \pm 0.05$ 
& $95.20 \pm 0.10$ \\
\midrule
\midrule
\multirow{2}{*}{Method} 
& \multicolumn{2}{c|}{OxfordPets} 
& \multicolumn{2}{c}{StanfordCars} \\
& Base & New 
& Base & New \\
\midrule
MMRL++ 
& $95.06 \pm {0.49}$ 
& $97.71 \pm 0.33$ 
& $81.38 \pm 0.54$ 
& $75.22\pm 0.09$ \\

\textbf{+MQAdapter}
& $96.70  \pm 0.32$ 
& $97.65 \pm 0.18$ 
& $82.08  \pm 0.32$ 
& $75.39  \pm 0.19$ \\
\midrule
\midrule
\multirow{2}{*}{Method} 
& \multicolumn{2}{c|}{Flowers102 } 
& \multicolumn{2}{c}{Food101} \\
& Base & New 
& Base & New \\
\midrule
MMRL++ 
& $97.72 \pm {0.16}$ 
& $77.33 \pm 0.83$ 
& $90.41 \pm 0.12$ 
& $91.69 \pm 0.08$  \\

\textbf{+MQAdapter}
& $98.67 \pm 0.15$ 
& $78.01 \pm {0.27}$ 
& $90.79 \pm {0.09}$ 
& $91.97 \pm {0.02}$ \\
\bottomrule
\end{tabular}
}
\end{table}

\subsection{Visualization Results Analysis}
We visualize the learned visual features with t-SNE on the base and new classes of EuroSAT and OxfordPets datasets. The results are demonstrated in Fig.~\ref{visual}. The CLIP provides coarse category-level separability, but still struggles to distinguish semantically similar classes. This observation is consistent with our motivation. Compared with CLIP, MMRL++ adapts the visual representation to downstream data, resulting in more compact class-wise clusters and improved inter-class separability. By incorporating MQAdapter, the learned features become more discriminative, leading to larger inter-class margins and more compact intra-class distributions. These results indicate that MQAdapter effectively leverages semantically related textual cues to refine visual representations. Moreover, MQAdapter models higher-order cross-modal interactions in a high-dimensional Hilbert space between visual and textual modalities. This enhances the visual encoder’s ability to capture fine-grained semantic features, thereby achieving better generalization on both base and new classes.

\subsection{Efficiency Analysis of MQAdapter}
To evaluate the efficiency of our MQAdapter, we report the number of learnable parameters and inference speed compared with other state-of-the-art methods. The results are shown in Tab.~\ref{efficy}, where FPS is measured with a batch size of 100 on a single NVIDIA RTX 3090 GPU. Compared with the MMRL++ framework, MQAdapter introduces only $0.078$M additional learnable parameters. Specifically, the proposed Variational Quantum Circuit (VQC) module requires only $0.024$K parameters, which is substantially fewer than a standard fully connected layer operating on conventional feature space. These results indicate that the proposed MQAdapter maintains competitive inference efficiency while consistently improving adaptation performance.

\begin{table}[t]
\centering
\caption{Computational efficiency of MQAdapter compared with other methods. ‘V-L’ denotes vision-language interaction in different layers. `L’ refers to fine-tuning textual modality alone.}
\setlength{\tabcolsep}{3pt}
\begin{tabular}{lcccccc}
\toprule
Method & Modality & Params & FPS (100) & Base & New & H \\
\midrule
TCP       & L   & 0.332M  & 543   & 84.13 & 75.36 & 79.51 \\
MMA       & V-L & 0.675M  &  363   & 83.20 & 76.80 & 79.87 \\
\midrule
MMRL++    & V-L & 0.813M  &  386   & 85.46 & 77.96 & 81.54 \\
\textbf{+MQAdapter} & V-L & +0.078M  & 300    & 86.09 & 78.93 & 82.36 \\
\bottomrule
\end{tabular}
\label{efficy}
\end{table}

\section{Conclusion}
In this paper, we introduced MQAdapter, a parameter-efficient coarse-to-fine adaptation framework for few-shot VLM fine-tuning. 
MQAdapter first retrieves semantically relevant Top-$K$ textual candidates from the frozen VLM as anchors, and then leverages them to guide fine-grained visual representation refinement. 
By mapping visual and textual features into a quantum-enhanced representation space, MQAdapter leverages quantum superposition and entanglement to capture higher-order visual-text interactions, enabling effective adaptation of frozen VLMs. 
Extensive experiments on four tasks demonstrate that MQAdapter consistently improves downstream performance while requiring fewer trainable parameters. 
Moreover, MQAdapter can be integrated with existing VLM fine-tuning methods, leading to further performance gains.

\bibliographystyle{IEEEtran}
\bibliography{reference}
\end{document}